\documentclass[final,conference,10pt]{IEEEtran}
\usepackage{caption}
\usepackage{amssymb}
\usepackage{amsmath}
\usepackage{amsthm}
\usepackage{mdframed}
\usepackage{graphicx}
\usepackage{xfrac}
\usepackage{booktabs}
\usepackage{color, colortbl}
\usepackage{enumitem}
\usepackage{subcaption}
\usepackage{algorithmic}
\usepackage{algorithm}

\usepackage{hyperref}
\definecolor{LightCyan}{rgb}{0.88,1,1}
\definecolor{Gray}{gray}{0.9}
\def\SumN{\sum_{t=1}^{T}}
\def\Sum{\sum_{i=1}^t}
\def\AA{\alpha}
\def\O{\mathcal{O}}

\def\DD{\Delta}
\def\dd{\delta}

\def\e{\varepsilon}

\def\ll{\lambda}
\def\LL{\Lambda}
\def\L{\mathcal L}

\def\RR{\mathbb R}

\def\X{\mathcal{X}}

\def\ts{\textstyle}

\def\c{\mathcal}
\def\lee{\preceq}
\def\grad{\nabla}
\def\p{\tilde}

\def\ol{\overline}
\theoremstyle{definition}

\newcounter{T} 
\newtheorem{theorem}[T]{Theorem}
\newcounter{L} 
\newtheorem{lemma}[L]{Lemma}

\newtheorem{corollary}{Corollary} 
\IEEEoverridecommandlockouts
\begin{document}

\title{Lazy Lagrangians with Predictions for Online Optimization}
\title{Lazy Lagrangians for Optimistic Learning}
\title{Lazy Lagrangians with Predictions for Online Learning}


\author{Daron Anderson, George Iosifidis, and Douglas J. Leith\thanks{{D. Anderson, D. Leith are with Trinity College Dublin, Ireland (AndersD3@tcd.ie, Doug.Leith@tcd.ie). G. Iosifidis is with Delft University of Technology, The Netherlands (G.Iosifidis@tudelft.nl; corresponding author).}}}

\maketitle
\begin{abstract} 
We consider the general problem of online convex optimization with time-varying additive constraints in the presence of  predictions for the next cost and constraint functions. A novel primal-dual algorithm is designed by combining a Follow-The-Regularized-Leader iteration with prediction-adaptive dynamic steps. The algorithm achieves $\c O(T^{\frac{3-\beta}{4}})$ regret and $\c O(T^{\frac{1+\beta}{2}})$  constraint violation bounds that are tunable via parameter $\beta\!\in\![1/2,1)$ and have constant factors that shrink with the predictions quality, achieving eventually $\c O(1)$ regret for perfect predictions. Our work extends the FTRL framework for this constrained OCO setting and outperforms the respective state-of-the-art greedy-based solutions, without imposing conditions on the quality of predictions, the cost functions or the geometry of constraints, beyond convexity.
\end{abstract}

\section{Introduction}
The online convex optimization (OCO)  framework introduced in \cite{gordon-colt99} and \cite{zinkevich} is employed to solve various learning problems ranging from spam filtering  to portfolio selection and network routing, cf.  \cite{hazan-book}. At each round $t$ an algorithm selects an action $x_t$ from a convex set $\X\subseteq \mathbb R^N$ and incurs cost $f_t(x_t)$, where the convex function $f_t:\X\mapsto \mathbb R$ is revealed after $x_t$ is decided. The algorithm's performance is measured using the metric of \emph{regret}:
\begin{equation}
\c R_T=\ts\SumN\big( f_t(x_t)-f_t(x^\star) \big), \label{regret}
\end{equation}
which quantifies the difference of the total cost from that of the best action selected with hindsight $x^\star\!\in\!\arg\min_{x\in \X} \sum_{t=1}^T f_t(x)$.   The goal is to select actions $\{x_t\}$ that ensure sublinear regret, i.e., $\c R_T=o(T)$.

A practical extension of this setting is the \emph{constrained} OCO framework, where the actions must satisfy long-term constraints of time-varying functions:
\begin{equation}
g_t(x)\triangleq\big(g_t^{(1)}(x), g_t^{(2)}(x), \ldots, g_t^{(d)}(x) \big)\lee 0, \notag
\end{equation}
which are unknown when $x_t$ is decided. In this case we are additionally interested in achieving sublinear total constraint violation,  $ \c V_T = o(T)$, where:
\begin{equation}
\c V_T = \Big \| \left[\ts \SumN  g_t(x_t)\right]_+ \Big\|. \label{const-viol}
\end{equation}

Constrained OCO algorithms have applications in  advertising with budget constraints, control of capacitated communication systems,  queuing problems \cite{neely-book}, etc. Nevertheless, they are notoriously hard to tackle. In particular, \cite{tsitsiklis} showed that no algorithm can achieve sublinear regret and constraint violation relative to each benchmark: 
\begin{equation}
x^\star \in \X_T^{\text{max}}=\Big\{x\in \X \, \Big\vert \, \ts \sum_{t=1}^T g_t(x)\lee 0 \Big\}. \notag
\end{equation}
Subsequent works considered either benchmarks that respect the constraints for short time windows \cite{paschos-icml}; dynamic benchmarks $\{x_t^\star\}$ that satisfy separately each $t$-round constraint $g_t(x_t^\star)\!\lee\! 0$ \cite{johansson-TSP2020}, \cite{giannakis-TSP17}; or benchmarks \cite{kapoor-icml17}, \cite{neely-nips17} restricted in set:
\begin{equation}
\X_T=\left\{x\in \X \, \Big\vert \, g_t(x)\lee 0,\,\forall t\leq T \right\}.  \notag
\end{equation}
Special cases of $\X_T$ are considered in \cite{mahdavi-jmlr2012,jennaton,lamperski-nips2018, bandit-knapsacks-2019} where $g_t(x)=g(x)$, $\forall t$; and in \cite{victor} which focuses on linearly-perturbed constraints $g_t(x)=g(x)+b_t$.

An aspect that has received less attention, however, is whether constrained OCO algorithms can be assisted by predictions for the next-round functions $f_t$ and $g_t$. Such information can be provided by a pre-trained model that uses incomplete data and hence cannot be fully trusted -- yet, can still assist the online algorithm. Leveraging predictions to improve learning algorithms is attracting increasing interest and has many practical applications, e.g., in data caching \cite{lykouris-icml2018}; online rent-or-buy problems \cite{kumar-neurIPS2018}; and in scheduling algorithms \cite{lattanzi-SODA2020}, among other areas. In this context, a key challenge is that the predictions might exhibit time-varying and unknown accuracy, which, furthermore, may vary across the cost and constraint functions. This confounds their incorporation in online learning algorithms and raises the question: \emph{how much can predictions improve the performance of constrained OCO algorithms and how can we accrue these benefits in the presence of inaccurate, potentially even adversarial, predictions?}

\subsection{Related Work}

Early works studying the impact of predictions include \cite{hazan-colt07} where the algorithm has access to the first coordinate of the cost vector; and \cite{dekel-nips17} which considered linear costs $c_t\!=\!\grad f_t(x_t)$ and predictions $\p c_t$ with guaranteed correlation $\p c_t^\top c_t\ge \AA \|c_t\|^2$. These predictions improve the regret from $\O(\sqrt T)$ to $\O(\log T)$. \cite{google-2020} considered the case when at most $B$ of the predictions fail the correlation condition and provided an $\O\big( (1\!+\!\sqrt B)/\AA)\log(1\!+T\!-B) \big)$ regret algorithm, that was further extended to combine multiple predictors \cite{google-nips-2020}. However, these prior models assume $\X$ is time-invariant. A different line of works \cite{sridharan-nips2013}, \cite{mohri-aistats2016} use adaptive regularizers and define prediction errors $\e_t=\|c_t - \p c_t\|$ to obtain $\O\big (\sqrt{\sum_{t} \e_t^2}\big)$ regret bounds. We adapt these methods to the time-varying constrained setting ($x_t\!\in\!\X_T$) where we incorporate predictions for the cost \emph{and} constraint vectors. Finally, \cite{wierman-1, wierman-2, wierman-3} consider a fixed $\X$ and multiple predictions over a time window, and bound the \emph{expected} performance or assume some special cost structure.


While the above works incorporate predictions in OCO problems with fixed time-invariant constraints (i.e., $x\!\in \X$), we focus on the richer constrained OCO setting where $\{x_t\}$ are also subject to \emph{time-varying} budget constraints ($x\in \X_T$) for an \emph{unknown} horizon $T$. We use a more general model than prior studies with no  assumptions on the predictions quality, the geometry of set $\X_T$ or the functions $\{f_t, g_t\}$, beyond being convex. Technically, our approach benefits from a novel ``lazy'' update that aggregates all previous cost and constraint vectors and uses \emph{data-driven} steps that adapt to prediction errors. In particular, we build on FTRL, cf. \cite{mcmahan-survey17}, which we extend here with time-varying accumulated constraints --- a result of independent interest. Previous greedy-based algorithms for time-varying budget constraints and benchmarks in $\X_T$ include \cite{kapoor-icml17, paschos-icml} which achieve $\c R_T\!=\!\O(\sqrt T)$ and $\c V_T\!=\!\O(T^{3/4})$ assuming, however, fixed and known horizon $T$; \cite{victor} that offers $\c R_T, \c V_T=\O(\sqrt T)$ but confines the constraints to be linearly-perturbed; \cite{neely-nips17} with $\c R_T, \c V_T=\O(\sqrt T)$ that restricts the constraints to be i.i.d. stochastic; and \cite{giannakis-TSP17} that supports $\c R_T, \c V_T=\O(T^{2/3})$ when $T$ is known and the constraints vary slowly; see also Table \ref{table-bounds}. Importantly, none of them can include and benefit from predictions.

\subsection{Contributions}

We study the general constrained OCO problem where in round $t$ our algorithm, which we name LLP (\emph{Lazy Lagrangians with Predictions}), has access to all prior cost gradients $\{\grad f_i(x_i)\}_{i=1}^{t-1}$ and constraints $\{ g_i(x)\}_{i=1}^{t-1}$, and receives predictions $\p g_t(\p x_t)$, $\grad \p f_t(\p x_t)$ and $\p g_t(\cdot)$. After selecting $x_t$, LLP incurs cost $f_t(x_t)$ and violation $g_t(x_t)$, and the process repeats in the next round. Our first result, Theorem \ref{Theorem1}, presents the regret and constraint violation bounds and demonstrates how they benefit from predictions. Theorem \ref{Theorem2} characterizes the (tunable) growth rates of the bounds and exhibits their dependency on the accumulated prediction errors. Theorem \ref{Theorem3} and Lemma \ref{lem:linearized} present the respective bounds when LLP employs fully-linearized cost and constraint functions and non-proximal regularizers, cf. \cite{mcmahan-survey17}, in order to reduce its computation and memory requirements. For this linearized version, it suffices to have gradient predictions $\grad \p g_t(\p x_t)$ instead of predictions $\p g_t(\cdot)$ for the entire constraint function. Indeed, in some problems it might be easier to acquire such single-point predictions compared to predictions for the constraint function; but we note that this is not always the case\footnote{For example, when $g_t(x)$ is the (non-linear) monetary cost for purchasing $x$ units of a resource, predicting $\p g_t(\cdot)$ requires knowing the price per unit; while $\grad \p g_t(\p x_t)$ requires \emph{also} to predict the actual purchased amount $\p x_t$. }; LLP can handle both scenarios. Finally, Lemma \ref{lem:linear-const} presents LLP's performance for linearly-perturbed constraints, a special but important case that was studied in \cite{victor}.

The performance of LLP is summarized in Table \ref{table-bounds}. LLP achieves $\c R_T\!=\c O(T^{\frac{3-\beta}{4}})$ and $\c V_T\!=\c O(T^{\frac{1+\beta}{2}})$ for worst-case (or, no) predictions, which are tunable through parameter $\beta\in[1/2,1)$. For instance, with $\beta\!=\!1/2$, we obtain  $\c R_T\O(T^{\frac{5}{8}})$ and, $\c V_T=\O(T^{\frac{3}{4}})$, that are further reduced to $\c R_T, \c V_T\!=\O(\sqrt T)$ when $-\c R_T\!=\O(T^{\frac{1}{2}})$, i.e., when $\{x_t\}$ does not outperform $x^\star$ by more than that. With perfect predictions, LLP achieves $\c R_T\!=\O(1)$, $\c V_T\!=\O(T^{\frac{1+\beta}{2}})$ which are tunable via $\beta\in[0,1)$; while for linearly-perturbed constraints (as in \cite{victor}) LLP ensures $\c R_T=\c O(\sqrt T)$ and $\c V_T=\O(T^{5/8})$. These results improve previously-known bounds for the general problem, i.e., without imposing additional assumptions such as strong convexity of functions and domains, or fixed $T$. Finally, they include as special cases the benchmarks with static or stationary constraints of \cite{mahdavi-jmlr2012, jennaton, lamperski-nips2018, neely-nips17, neely-JML2020}. Importantly, unlike \emph{all prior} constrained-OCO algorithms, the constant factors of $\c R_T$ and $\c V_T$ shrink proportionally to the predictions' accuracy, an advantage that is revealed even with simple numerical examples.

\begin{table}[t]
	\begin{small}
		\centering
		\begin{tabular}{ c c c c c } 
			\textbf{Paper} & 		\textbf{Regret} & 		\textbf{Const.} & \textbf{Conditions}\\ 
			\toprule[0.9pt]
			\noalign{\smallskip}
			\!\!\!\!\!\cite{johansson-TSP2020}	& $\O(\sqrt T)$	 &	$\O(\sqrt T)$	& \footnotesize{Slater; strongly cvx $f_t, g_t$} \\
			\noalign{\smallskip}
			\!\!\!\!\!\cite{victor}		 & $\O(\sqrt T)$	&	$\O(\sqrt T)$  &  Slater; $g_t(x)\!=g(x)\!+\! b_t$	\\ 
			\noalign{\smallskip}
			\!\!\!\!\!\cite{kapoor-icml17}		& $\O(\sqrt T)$	&		$\O(T^{\frac{3}{4}})$ &   $L_f,L_g,F, G, T$ known\\ 
			\noalign{\smallskip}
			\!\!\!\!\!\cite{giannakis-TSP17}	 & $\O(T^{\frac{2}{3}})$	 &	$\O(T^{\frac{2}{3}})$	& \footnotesize{$\|g_t(x)\!-\!g_{t-1}(x)\|\!\leq\! \epsilon$; $T$ known} \\  
			\noalign{\smallskip}
			\!\!\!\!\!\cite{paschos-icml}	& $\O(\sqrt T)$ &		$\O(T^{\frac{3}{4}})$ & Slater; $T$ known 	\\ 
			\noalign{\smallskip}
			\hline
			\noalign{\smallskip}
			\rowcolor{Gray}
			LLP			 & $\O(T^{ \frac{3-\beta}{4} })$		 &		$\O(T^{ \frac{1+\beta}{2} })$ & no predictions; $\beta\in[\frac{1}{2},1)$	\\ 
			\rowcolor{Gray}			
			&	  $\O(\sqrt T)$		 &		$\O(\sqrt T)$ &  $-\c R_T \!=\! \O(\sqrt T)$ 	\\ 
			\rowcolor{Gray}			
			&	 	 $\O(1)$		 &		$\O(T^{\frac{1+\beta}{2} })$ &  perfect predictions; $\beta\in[0,1)$	\\  			
			\rowcolor{Gray}			
			&	 	 $\O(\sqrt T)$		 &		$\O(T^{ \frac{5}{8} })$ &  no predict.; $g_t(x)\!=g(x)\!+\! b_t$	\\  						
			\bottomrule[0.99pt]
			\noalign{\smallskip}
		\end{tabular}
		\caption{Point $x^\star$ belongs in  $\X$ and satisfies $g_t(x^\star)\preceq 0, \forall t$. For the algorithms with tunable bounds we present the best achievable w.r.t. $\c R_T$.}. 
		\label{table-bounds}
	\end{small}
\end{table}

\vspace{-1mm}
\subsection{Assumptions and Notation}
We write $\{x_t\}$ for a sequence of vectors and use subscripts to index them; $\|\cdot\|$ denotes the Euclidean ($\ell_2$) norm and $[x]_{\X}, [x]_+$ the $\ell_2$-projection of $x$ on sets $\X$ and $\mathbb R_+^N$. We use the index function $\mathbf{I}_{\X}(x)\!=\!0$ if $x\!\in\!\X$ and $\mathbf{I}_{\X}\!=\!\infty$, otherwise. Vector $c_t$ denotes the gradient $\grad f_t(x_t)$ of $f_t$ or an element of its subdifferential $\vartheta f_t(x_t)$ if it is non-differentiable; and $\grad g_t(x)$ denotes the Jacobian of the vector-valued constraint. We use the shorthand notation $c_{1:t}$ for $\sum_{i=1}^t c_i$, and $\p a_t$ for the prediction of some vector (or,  function) $a_t$.

The analysis requires the following basic assumptions.

\noindent \textbf{A1}. The set $\mathcal X\!\subset \RR^N$ is convex and compact, and it holds $\|x\| \leq D$, $\forall x\in\X$. 

\noindent \textbf{A2}. Functions $f_t, g_t^{(j)}\!:\! \mathcal X\!\mapsto \!\mathbb R$, $\forall t, j\!\leq\!d$, are convex and Lipschitz with constants $L_{f_t}\!\leq\! L_f$, $L_{g_t^{j}}\!\leq\! L_g$. Since $\X$ is compact it holds $|f_t(x)|\!\leq\! F$, $\|g_t(x)\|\!\leq\! G$, $\forall t, x\!\in\!\X$.

\noindent \textbf{A3}. Predictions $\p c_t$, $\p g_t(\cdot)$ and $\p g_t(\p x_{t})$ are known at $t$.

\noindent \textbf{A4}. The prediction errors $\e_t\!\triangleq\!c_t\!-\p c_t$, $\dd_t\!\triangleq\!\grad g_t(x_t)- \grad \p g_t(x_t)$ are bounded: $\|\e_t\|\!\le\!E_{m}$, $\|\dd_t\|\!\le\!\Delta_m$, $\forall x_t\in\X$; and it holds $\|\p c_t\|\leq L_f$, $\|\p g_t(x)\|\leq G$, $\forall t, x\in\X$.

\subsection{Paper Organization}

Section \ref{sec:llp} introduces the LLP algorithm and the regret and constraint violation bounds. Section \ref{sec:rates} presents the adaptive multi-step and characterizes the convergence rate of LLP, with special focus to the case of perfect predictions and worst-case (or no) predictions. Section \ref{sec:less} modifies LLP for linearized constraints and non-proximal updates, and Sec. \ref{sec:pert} derives the performance bounds for the special case of linearly-perturbed constraints. We conclude in Sec. \ref{sec:conclusions}. The paper is accompanied by an appendix, Sec. \ref{sec:appendix}, that includes the remaining proofs, explanatory figures, and numerical examples. 

\section{The LLP Algorithm} \label{sec:llp}

Our approach is inspired by saddle-point methods that perform min-max operations on a convex-concave Lagrangian. Starting from the $t$-round problem:
\begin{equation}
\min_{x\in \X}\,\,f_t(x)\quad \text{s.t.} \quad g_t(x)\lee 0, \notag
\end{equation}
we introduce the dual variables $\ll\!\in\! \mathbb R_+^d$ by relaxing $g_t(x)\!\preceq\! 0$, and define the \emph{regularized} Lagrangian:
\begin{align}
\L_t(x,\ll)=r_t(x) + c_t^\top x+\ll^\top g_t(x)-q_t(\ll), \label{eq:lagrangian} 
\end{align} 
where we linearized $f_t(x)$. Function  $r_t\!:\!\X\!\mapsto\! \mathbb R$ is a proximal\footnote{Regularizer $r_t(x)$ is called \emph{proximal} with reference to an algorithm that yields $\{x_t\}$ if $x_t\!\in\!\arg\min_{x\in\X} r_{t}(x)$; and  \emph{non-proximal} otherwise \cite{mcmahan-survey17}.} primal regularizer and $q_t:\!\mathbb R_+^d\!\mapsto\! \mathbb R$ a non-proximal dual regularizer. We also set $\L_0(x,\ll)=r_0(x)\!-q_0(\ll)$.

\begin{figure}[t]
	\centering
	{\includegraphics[width=0.95\columnwidth]{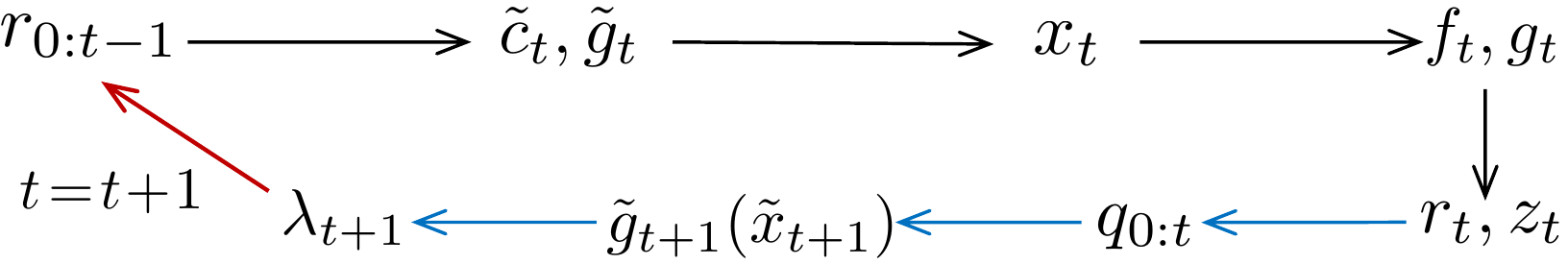}}
	\caption{{\small{Key steps, timing and predictions of LLP.}}}
	\label{Fig:llp}
\end{figure}

We coin the term \emph{Lazy Lagrangians with Predictions} (LLP) for Algorithm \ref{alg:llp}, which proceeds as follows, cf. Fig. \ref{Fig:llp}. In each round $t$, LLP uses observations $\{c_i\}_{i=1}^{t-1}$, $\{g_i\}_{i=1}^{t-1}$, dual variables $\{\ll_i\}_{i=1}^t$, and predictions $\p c_t$, $\p g_t(\cdot)$ to perform an \emph{optimistic} FTRL update:
\begin{equation}
x_t\! =\! \arg \min_{x\in \X} \ts  \left\{ \sum_{i=0}^{t-1} \L_i(x,\ll_i) +\p c_t^\top x +\ll_t^\top \p g_t(x) 	\right\}, \! \label{primal-update}
\end{equation}
which induces cost $f_t(x_t)$ and constraint violation $g_t(x_t)$. After the $t$-round information $f_t(x)$ and $g_t(x)$ is revealed, LLP calculates the  \emph{prescient} action:
\begin{equation}
  z_t = \arg \min_{x\in \X} \ts \left\{ \sum_{i=0}^{t} \L_i(x,\ll_i)\right\}, \label{prescient-update}
\end{equation}
and uses prediction  $\p g_{t+1}(\p x_{t+1})$ to update:
\begin{equation}
\ll_{t+1}\!=\! \arg \max_{\ll\!\in\mathbb R_+^d} \ts \left\{ \sum_{i=0}^t \L_i(z_i,\ll)\! + \ll^\top \p g_{t+1}(\p x_{t+1})\right\} \! \label{dual-update}
\end{equation}
where note the use of $\{z_t\}$ instead of $\{x_t\}$. The process then repeats in the next round.

\begin{algorithm}[t]
	\caption{{{Lazy Lagrangians with Predictions (\textbf{LLP})}}}
	\label{alg:llp}
	\begin{algorithmic}
		\STATE {\bfseries Input:}  $x_0 \!\in\! \X$,  $\ll_1\!=\!0$, $r_t(x)$, $q_t(\ll)$ with \eqref{primal-reg}, \eqref{dual-reg}.
		\FOR{$t=1,2,\ldots$  }  
		\STATE $\circ$ Calculate $r_{0:t-1}$ and decide $x_t$ using \eqref{primal-update}\;	
		\STATE $\bullet$ Pay cost $f_t(x_t)$ and violation $g_t(x_t)$\; 
		\STATE $\circ$ Calculate $r_{0:t}$ and decide $z_t$ using \eqref{prescient-update}\;
		\STATE $\bullet$ Receive predictions $\p c_{t+1}$, $\p g_{t+1}(\cdot)$, $\p g_{t+1}(\p x_{t+1})$\;
		\STATE $\circ$ Calculate $q_{0:t}$ and decide $\ll_{t+1}$ using \eqref{dual-update}\;
		\ENDFOR   
	\end{algorithmic} 
\end{algorithm}

LLP has key differences from previous constrained OCO algorithms. These stem from the usage of \emph{lazy} as opposed to \emph{greedy} updates in the primal and dual iteration, where instead of using $x_{t-1}$ and $\ll_t$ to decide $x_t$ and $\ll_{t+1}$ respectively, we aggregate in a projection-free fashion all prior cost gradients and constraints. This approach can be traced back to lazy algorithms discussed in \cite{zinkevich}; to \emph{fictitious} (as opposed to \emph{best response}) strategies in game theory \cite{levin-learning}; and to FTRL algorithms  \cite{mcmahan-survey17} for problems with fixed constraints. However, to the best of our knowledge, this is the first time such lazy updates are used with time-varying budget constraints.

\textbf{Performance}. The regret and constraint violation are quantified using \eqref{regret} and \eqref{const-viol}, respectively, with benchmark $x^\star\!\in\!\arg\min_{x\in \X_T}\SumN f_t(x)$. The performance of LLP is shaped by the regularizers which adapt to predictions. In particular, we use the primal regularizers:
\begin{align}
&\!r_0(x)\!=\!\mathbf{I}_{\X}(x)\,\,\,\text{and}\,\,\, r_t(x)\!=\!\sigma_t\|x\!-x_t\|^2/2,\,t\geq 1\,\,\,\text{with:} \notag\\	
&\!\sigma_t\!=\!\sigma\! \left(\!\! \sqrt{ h_{1:t} }\! -\! \sqrt{h_{1:t-1} }  \right),\,\, h_t\!\triangleq\!\|\e_t\!+\!\ll_t^\top\dd_t\|,\,\,\sigma\!>\!0, \label{primal-reg}
\end{align}
where $r_0(x)$ ensures that $x\!\in \X$; $x_t$ is given by \eqref{primal-update}; and the regularization parameter $\sigma_t$ accounts for the cost and constraint prediction errors, where the latter are modulated by the dual variables. The intuition for \eqref{primal-reg} is that we  add regularization commensurate to the prediction errors; and  the rationale for selecting this particular $\sigma_t$ will be made clear below. On the other hand, we use the general dual regularizer:
\begin{align}
&\! q_0(\ll)=\mathbf{I}_{\mathbb{R}_+^d}(\ll)\,\,\,\,\text{and}\,\,\,\, q_t(\ll)=\phi_t\|\ll\|^2/2,\,\,t\geq1\,\,\,\text{with:}\notag \\ 
&\phi_t\!=(1/a_t)-(1/a_{t-1}), \,\,a_{t-1}\geq a_t>0,\,\,\phi_0\!=\!1/a_0 \!\!\!\label{dual-reg}
\end{align}
where again $q_0(\ll)$ ensures $\ll\!\in\!\mathbb R_+^d$, and $\{a_t\}$ is the dual learning rate \cite{mcmahan-survey17} which, for the first Theorem, suffices to be non-increasing.  Our first main result is the following. 
\begin{mdframed}[innerrightmargin=5pt,innerleftmargin=5pt] \vspace{1mm}
\begin{theorem}\label{Theorem1}
Under Assumptions (\textbf{A1})-(\textbf{A4}) and with $\{r_t\}$ and $\{q_t\}$ satisfying \eqref{primal-reg} and \eqref{dual-reg}, LLP ensures $\forall x^\star\in\X_T$: 
\begin{align} 
&\!\!\!\!\!\mathbf{(a)}\!: \c R_T \le B_T\triangleq 2\Big(\sigma D^2 \!+\!\frac{L_f}{\sigma}\Big)\sqrt{h_{1:T}} + \sum_{t=1}^T \frac{\xi_t^2}{\phi_{0:t-1}}   \notag \\  
&\!\!\!\!\!\mathbf{(b)}:\c V_T\! \le \! \sqrt{2\phi_{0:T-1}(B_T-\!\c R_T)}+\frac{2L_g}{\sigma}\sqrt{h_{1:T}} \notag 
\end{align}
where $\xi_t\triangleq\|g_t(z_t)-\p g_t(\p x_t)\|$, $h_t\triangleq \| \e_t+\ll_t^\top \dd_t\|$, $ \e_t\triangleq\!c_t-\p c_t$, $\dd_t\!\triangleq\!\grad g_t(x_t)- \grad \p g_t(x_t)$.
\end{theorem}
\end{mdframed}

\textbf{Discussion}. We can see from Theorem \ref{Theorem1} the effect of predictions on the bounds of $\c R_T$ and $\c V_T$, which diminish proportionally to their accuracy. The bound of $\c R_T$ is further reduced when $\sigma\!=\!\sqrt{L_f}/D$ and it settles to zero for perfect predictions, i.e. when $x_t=z_t, \forall t$, and:
\[
\epsilon_t=0,\,\,\,\,\,\delta_t=0,\,\,\,\,\,\xi_t=0,\,\,\,\forall t\leq T,
\]
while the same is not true for $\c V_T$. Moreover, this theorem reveals the \emph{tension} between $\c R_T$ and $\c V_T$. Indeed, observe that $-\c R_T$ appears in the bound of $\c V_T$ which means that when $\{x_t\}$ outperforms $x^\star$, we might incur higher constraint violation. 

From Algorithm \ref{alg:llp} we can see that LLP requires predictions for the next gradient $\p c_{t}\!=\!\grad\p f_t(\p x_t)$, next cost function $\p g_t(\cdot)$, and next constraint point $\p g_t(\p x_t)$. It is important here to note the timing of these predictions. Namely, updating $x_t$ requires $\p c_t$ and $\p g_t(\cdot)$ and access to regularizers $r_{0:t-1}$ which are calculated using the prediction errors up to slot $t-1$.  Knowledge of $\p c_t$ is the standard prediction that all prior works employ, e.g., see \cite{google-nips-2020}, \cite{dekel-nips17} and references therein. On the other hand, since we have not linearized the constraint function, the respective predictions involve function $\p g_t(\cdot)$ and its next-round value $\p g_t(\p x_t)$. In Section \ref{sec:less} we present a version of LLP where we linearize the constraints and hence use only gradient predictions for the constraints -- similarly to cost functions.

The complexity of LLP is analogous to its greedy-based counterparts --- sans the additional prescient update \eqref{prescient-update} --- i.e., it requires the solution of strongly convex problems and a closed-form iteration for the dual update; the calculation of the regularizers and steps is in $\O(1)$. Finally, it is worth emphasizing that the impossibility result of \cite{tsitsiklis}  holds even if $\{f_t, g_t\}$ are revealed before $\{x_t\}$ is selected, as stated in the next Lemma that is proved in the Appendix. This exhibits the challenges in tackling constrained OCO problems. 
\begin{lemma}\label{lem:impossibility}
No online algorithm can achieve concurrently sublinear regret and constraint violation, $\c R_T=o(T)$, $\c V_T=o(T)$, w.r.t. $x^\star \in \X_T^{\text{max}}$, even if the algorithm selects $\{x_t\}$ with knowledge of $\{f_t, g_t\}$.
\end{lemma}
\noindent The next subsections prove the $\c R_T$ and $\c V_T$ bounds. 
 
\subsection{Regret Bound}\label{sec:regret}
Our strategy is to derive a regret bound w.r.t. prescient actions $\{z_t\}$, and then use the distance of $\{z_t\}$ from $\{x_t\}$ to prove Theorem 1\textbf{(a)}. We will use the following Lemma that is proved in the Appendix. 

\begin{lemma}\label{2ssclaim2b}
For the actions $\{x_t\}$ and $\{z_t\}$ obtained by \eqref{primal-update} and \eqref{prescient-update}, respectively, it holds:
\begin{align}
\|x_t-z_t\| \leq \frac{\|\e_t+ \ll_t^\top\dd_t\|}{\sigma_{1:t}}, \notag   
\end{align}
where $\epsilon_t\!=\!c_t\!-\!\p c_t$, $\delta_t\!=\!\grad g_t(x_t)\!-\!\grad \p g_t( x_t)$, $\sigma_t$ from \eqref{primal-reg}.
\end{lemma}

Now, to prove Theorem \ref{Theorem1}\textbf{(a)} we apply \cite[Theorem 2]{mohri-aistats2016} to update\footnote{Update \eqref{dual-update} runs over the unbounded set $\mathbb R_+^d$, unlike the compact set in \cite{mohri-aistats2016}. However, that result still holds here and suffices as we set $\lambda\!=\!0$ to get $q_{0:T-1}(0)=0$; see discussion and Lemma \ref{predictionsb} in Appendix.} \eqref{dual-update} with functions $\{-\lambda g_t(z_t)\}$ and gradient predictions $\{-\p g_{t+1}(\p x_{t+1})\}$, to get:
\begin{align}
&-\sum_{t=1}^T \ll_t^\top g_t(z_t)+ \sum_{t=1}^T \ll^\top g_t(z_t) \leq \notag \\  
&q_{0:T-1}(\ll)+ \sum_{t=1}^T \|g_t(z_t)-\p g_t(\p x_t)\|_{(t-1),\star}^2 \stackrel{(a)}= \notag \\
&q_{0:T-1}(\ll)+ \sum_{t=1}^T \frac{\|g_t(z_t)-\p g_t(\p x_t)\|^2}{\phi_{0:t-1}},\,\,\,\forall \ll\in\mathbb R_+^d, \label{eq:dual1}
\end{align}
where $(a)$ holds since the  dual regularizer is 1-strongly-convex w.r.t. $\|\ll\|_{(t-1)}\!=\!\sqrt{\phi_{0:t-1}}\|\ll\|$ which has dual norm $\|\ll\|_{(t-1), \star}\!=\!\|\ll\|/ \sqrt{\phi_{0:t-1}}$.

Adding $\SumN\big[r_{t}(z_t) \!+\!c_t^\top z_t\big]$ to both sides of ineq. \eqref{eq:dual1}:
\begin{align}
&\SumN \Big[r_t(z_t) + c_t^\top z_t-\ll_t^\top g_t(z_t)+\ll^\top g_t(z_t)\Big]\leq\notag \\ 
&q_{0:T-1}(\ll) +\SumN \Big[ r_t(z_t) + c_t^\top z_t+ \xi_t^2/\phi_{0:t-1}\Big], \label{eq2}
\end{align}
and dropping the first non-negative sum in the LHS, setting $\ll\!=\!0$ to get $q_{0:T-1}(0)\!=\!0$, and adding/subtracting  $\phi_t \|\ll_t\|^2/2$ in the RHS so as to build $\L_t(z_t, \ll_t), \forall t\!\leq\!T$, we arrive at:  
\begin{align}
\SumN c_t^\top z_t &\leq \SumN \left( \L_t(z_t, \ll_t) +\frac{\phi_t\|\ll_t\|^2}{2}+ \frac{\xi_t^2}{\phi_{0:t-1}}\right)\notag \\
&\stackrel{(a)}\leq \SumN \left( \L_t(x^\star, \ll_t) +\frac{\phi_t\|\ll_t\|^2}{2}+ \frac{\xi_t^2}{\phi_{0:t-1}}\right) \notag \\
&\stackrel{(b)}\leq 2D^2\sigma_{1:T}+\SumN \left( c_t^\top x^\star+ \frac{\xi_t^2}{\phi_{0:t-1}}\right), \notag
\end{align}
where $(a)$ stems from the Be-the-Leader (BTL) Lemma \cite[Lemma 3.1]{CBGames} applied with $x^\star \!\in\! \X_T\!\subseteq\! \X$ to \eqref{prescient-update}; and $(b)$ from expanding $\L_t(x^\star, \ll_t)$, using $g_t(x^\star) \!\preceq 0$ and $\|x^\star\!-x_t\|^2\!\leq\! 4D^2, \forall t$. Add $\SumN c_t^\top x_t$ to both sides and rearrange: 
\begin{align}
\!\!\!\!\SumN\! c_t^\top (x_t\!-\!x^\star)\!\leq\! 2D^2\sigma_{1:T}\!+\!\SumN\!\frac{\xi_t^2}{\phi_{0:t-1}}\!+ c_t^\top\big(x_t\!-\!z_t\big)\!\!\label{eq1}
\end{align}
The last term can be upper-bounded using the Cauchy-Schwarz inequality and Lemma \ref{2ssclaim2b}, i.e.:
\begin{align*}
&\!\!\SumN c_t^\top\big(x_t\!-\!z_t\big)\!\leq\! \SumN \frac{\|c_t\||\e_t\!+\ll_t^\top\dd_t\|}{\sigma_{1:t}} \stackrel{(\text{Assump.}\,\,\textbf{A2})}\leq\! \\ 
&\!\!\frac{L_f}{\sigma}\SumN \frac{\|\e_t\!+\ll_t^\top\dd_t\|}{\sqrt{\Sum \|\e_i\!+\ll_i^\top\dd_i\|}}\! \stackrel{(a)}\leq\! \frac{2L_f}{\sigma}\sqrt{ \SumN \|\e_t+\ll_t^\top \dd_t\|}
\end{align*}
where in $(a)$ we used \cite[Lemma 3.5]{auer2002}, and this was made possible due to the specific formula of the regularization parameter  $\sigma_t$. Replacing in \eqref{eq1} and using $\c R_T\leq c_t^\top (x_t\!-x^\star)$, we eventually get:
\begin{align}
&\c R_T\leq\! 2D^2\sigma_{1:T}  +\! \frac{2L_f}{\sigma}\sqrt{ \SumN \|\e_t+\ll_t^\top \dd_t\|}+ \SumN  \frac{\xi_t^2}{\phi_{0:t-1}} \Rightarrow \notag\\
&\c R_T=\! 2\left(\sigma D^2\!+\!\frac{L_f}{\sigma}\right) \sqrt{ h_{1:T}}+ \SumN  \frac{\xi_t^2}{\phi_{0:t-1}} ,\notag
\end{align}
which concludes the proof for Theorem \ref{Theorem1}\textbf{(a)}.

\subsection{Constraint Violation Bound} \label{sec:constraint}
To prove Theorem \ref{Theorem1}\textbf{(b)}, we start from \eqref{eq2} where we drop again the non-negative $\SumN r_t(z_t)$ in the LHS, add and subtract the term $\SumN \phi_t\|\ll_t\|^2/2$, $\forall t$, and rearrange to get:
\begin{align*}
&\lambda^\top \SumN g_t(z_t) - q_{0:T-1}(\ll) \\
&\!\leq\!\SumN \left( \L_t(z_t,\ll_t)-c_t^\top z_t + \frac{\phi_t\|\ll_t\|^2}{2} + \frac{\xi_t^2}{\phi_{0:t-1}} \right)\\
&\!\leq\!\SumN \left( \L_t(x^\star,\ll_t)-c_t^\top z_t + \frac{\phi_t\|\ll_t\|^2}{2} + \frac{\xi_t^2}{\phi_{0:t-1}}\right)
\end{align*}
where we applied again BTL to $\L_t(z_t, \ll_t)$. Expand $\L_t(x^\star, \ll_t)$, use $q_{0:T-1}(\ll)\!=\!\phi_{0:T-1}\|\ll\|^2/2$, $g_t(x^\star)\!\preceq\!0, \forall t$,  and $r_{0:T}(x^\star)\!\leq 4D^2\sigma_{1:T}$, to get:
\begin{align}
&\lambda^\top \SumN g_t(z_t) - \frac{\phi_{0:T-1}}{2}\|\ll\|^2 \notag \\
&\leq2D^2\sigma_{1:T} \!+\!\SumN\left( c_t^\top(x^\star - z_t) + \frac{\xi_t^2}{\phi_{0:t-1}}\right) \notag \\
&\leq2D^2\sigma_{1:T} \!+\!  \SumN\! \left( c_t^\top(x^\star\!-\!x_t)  \!+\ c_t^\top(x_t \!- z_t) \!+ \frac{\xi_t^2}{\phi_{0:t-1}} \!\right)\notag \\
&\!\leq\!2D^2\sigma_{1:T} \!+ \frac{L_f}{\sigma}\sqrt{\SumN \|\e_t\!+\ll_t^\top\dd_t\|} \!+\SumN \frac{\xi_t^2}{\phi_{0:t-1}} \!-R_T \!\Rightarrow \notag \\
&\lambda^\top \SumN g_t(z_t)\!-\! \frac{\phi_{0:T-1}}{2}\|\ll\|^2\!\leq\! B_T\!-\c R_T,\,\,\forall \ll\in\mathbb{R}_+^d. \label{eq3}
\end{align}
For the LHS of \eqref{eq3}, we can use the result:
\begin{equation}
\frac{\left\| \left[ \SumN g_t(z_t) \right]_+\right\|^2}{2\phi_{0:T-1}}=\! \sup_{\ll\in\mathbb R_+}\!\left[\lambda^\top \SumN g_t(z_t) \!-\! \frac{\phi_{0:T-1}}{2}\|\ll\|^2\!\right] \notag
\end{equation}
and if we denote with $V_T^z$ the LHS norm and replace this in \eqref{eq3}, we obtain:
\begin{equation}
\!V_T^z\!\leq\! \sqrt{ 2\phi_{0:T-1}	\big(B_T\!-\c R_T\big)}\! =\! \sqrt{\frac{2(B_T\!-\c R_T)}{a_{T-1}}}.\!\!\!\label{vtz-bound}
\end{equation}
Lastly, we define $w_t\triangleq\grad g_t(x_t)^\top (z_t-x_t)$ and write: 
\begin{align}
	&\c V_T\!=\left\| \left[ \SumN g_t(x_t) \!+w_t-w_t \right]_+ \right\| \notag \\ 
	&\!\!\!\stackrel{(a)}\leq \left\| \left[ \SumN g_t(x_t) +w_t\right]_+\right\|+ \left\|\left[-w_{1:T}  \right]_+ \right\| \notag \\
	&\!\!\!\stackrel{(b)}\leq\!  \left\| \left[ \SumN g_t(z_t)\!\right]_+\right\| \!+\! \left\|-\!w_{1:T}  \right\| \stackrel{(c)}\leq\! V_T^z \!+\! \frac{2L_g}{\sigma} \sqrt{ h_{1:T} }\!\!\label{vtz-bound2}
\end{align}
where $(a)$ uses the identity $\| [\chi+\upsilon]_+\|\!\leq\! \|[\chi]_+\|\! +\! \|[\upsilon]_+\|$; $(b)$ the convexity of $g_t$; and $(c)$ the Cauchy-Schwarz inequality, $\|\grad g_t(x_t)\|\!\leq \!G$, and Lemma \ref{2ssclaim2b}.

The next section characterizes the convergence rates of the bounds, focusing on two special cases: when the predictions are perfect and when we have worst-case (or no) predictions.

\section{Convergence Rates} \label{sec:rates}

We start by specifying the dual learning rate $\{a_t\}$. The rationale for selecting the primal regularizer was made clear in the proof of Theorem \ref{Theorem1}; here, we refine \eqref{dual-reg} in a way that ensures the desirable sublinear regret and constraint violation growth rates. In detail, we will be using:
\begin{align}
a_t=\frac{a}{ \max\left\{ \sqrt{ 4G^2 + \sum_{i=1}^t \xi_i^2},\,\, t^\beta\right\}},\,\,\,\,\,\beta\in[0,1). \label{dual-rate}
\end{align}
This \emph{multi-step} combines the typical time-adaptive step appearing in online gradient-descent algorithms \cite{zinkevich} with a data-adaptive step that accounts for the prediction errors. This ensures that $a_t$ will induce enough regularization when the predictions are not satisfactory, but will also continue diminishing even in the case of perfect predictions -- a condition that is necessary in order to tame the growth rate of the dual vector. The additional term $4G^2\geq \xi_t^2, \forall t$ corrects the off-by-one regularizer of the non-proximal dual update.  

Before we analyze the convergence for the cases of perfect and worst predictions, it is important to emphasize that in each round $t$, LLP has at its disposal all the necessary information to calculate $a_t$. In particular, $a_t$ is used to update the dual vector $\lambda_{t+1}$ \emph{after} the cost and constraint functions, $f_t$ and $g_t$, have been revealed, and the prescient vector $z_t$ is calculated. Hence, we know $\xi_t$ before performing update \eqref{dual-update}.

\subsection{Perfect Predictions}

The next Corollary to Theorem \ref{Theorem1} describes the regret and constraint violation bounds for perfect predictions.
\begin{corollary}\label{corol:perfect}[Perfect Predictions]
	When  $\epsilon_t=\delta_t=\xi_t=0,\forall t$, $\beta\in[0,1)$, Algorithm LLP ensures:
	\[
	\c 	R_T=\O(1),\qquad \c V_T= \O(T^{\frac{1+\beta}{2}})
	\]	
\end{corollary}
Indeed, for perfect predictions it holds $h_{1:T}\!=\!0$ independently of the value of $\{\lambda_t\}$, while the second term in the bound of $\c R_T$ can be written (detailed derivation in Sec. \ref{sec:Th2proof}:
\[
\sum_{t=1}^T \frac{\xi_t^2}{\phi_{0:t-1}}=\sum_{t=1}^T a_{t-1}\xi_t^2=\min\left\{ 2a\sqrt{\sum_{t=1}^T\xi_t^2}, \frac{4aG^2}{1-\beta}T^{1-\beta} \right\}
\]
which diminishes to zero when $\xi_t=0, \forall t$. This manifests the advantage of this doubly-adaptive dual step which creates a bound similar to those in OCO problems without budget constraints, see \cite{mohri-aistats2016}. Furthermore, the step is simplified to $a_t\!=\!a/t^\beta$ which remains bounded. Hence $\c R_T\!=\!B_T\!=\!\O(1)$, and if we substitute $B_T$ in $\c V_T$, we get:
\begin{align}
	\c 	V_T\leq \sqrt{\frac{-2\c R_T}{a_{T-1}}}\leq \sqrt{ \frac{-2\c R_TT^{\beta}}{a} } = \O\left(T^{\frac{1+\beta}{2}}\right) \label{eq11}
\end{align}
Here, we can set $\beta\!=\!0$ to get $\c V_T\!=\O(T^{\frac{1}{2}})$, and reduce the constant factor of $\c V_T$ further by increasing\footnote{Had we known the horizon $T$, as assumed in \cite{kapoor-icml17}, we can set $a\!=T$ to obtain $\c V_T\!=\O(1)$. However, this selection endangers increasing $\c R_T$ when predictions are imperfect.} $a$. 

Finally, note that when the regret is non-negative, i.e., when the sequence $\{x_t\}_t$ does not outperform $x^\star$, then we get $\c V_T\leq 0$ for any value of $\beta$. And, more generally, if there is a non-trivial bound for the negative regret, i.e., $-\c R_T=\c O(T^b)$ with $b<1$, then the bound of the constraint violation is improved in a commensurate amount, namely $\c V_T=\c O\big(T^{ \frac{b+\beta}{2} }\big)$.

\subsection{Worst-Case Predictions}

On the other hand, when we do not have any predictions at our disposal, or when these are as far as possible from the actual data, then the dual multi-step induces more regularization using the observed prediction errors. The performance of LLP in this scenario is captured by the following theorem.

\begin{mdframed}[innerrightmargin=4pt,innerleftmargin=5pt, userdefinedwidth=23.6em] 
\begin{theorem}\label{Theorem2}
Under Assumptions (\textbf{A1})-(\textbf{A4}), with regularizers $\{r_t\}$, $\{q_t\}$ satisfying \eqref{primal-reg}, \eqref{dual-reg}, \eqref{dual-rate}, and for worst-case predictions $\epsilon_t\!=E_m, \delta_t\!=\Delta_m, \xi_t\!=2G, \forall t$, LLP ensures:
\begin{align*}
&\c R_T=\O\big(T^{\frac{5}{8}}\big),\qquad\,\,\,\,\, \c V_T= \O\big(T^{\frac{3}{4}}\big) \quad\,\,\,\,\, \text{when}\quad \beta\!<\!1/2 \\
&\c R_T=\O\big(T^{ \frac{3-\beta}{4}}\big),\qquad \c V_T= \O\big(T^{ \frac{1+\beta}{2} }\big) \,\,\,\,\, \text{when}\quad \beta\!\geq\!1/2
\end{align*}
\end{theorem}  
\end{mdframed}

We see that the growth rates are tunable by parameter $\beta\!\in\![0,1)$. For example, by setting $\beta\!=\!2/3$ we obtain $\c R_T\!=\!\O(T^{7/12})$ and $\c V_T\!=\!\O(T^{5/6})$; while with $\beta\!=3/5$ it is $\c R_T\!=\!\O(T^{3/5})$ and $\c V_T\!=\!\O(T^{4/5})$. These bounds improve the best-known results for the \emph{general} constrained-OCO problem, while being comparable with results that consider special cases, such as knowing the (a priori fixed) time horizon $T$ \cite{kapoor-icml17} or having only linearly-perturbed constraints \cite{victor}.

What prevents the LLP bounds from improving further is the term $-\c R_T $ that appears in $\c V_T$. While we have used in the analysis the worst-case $-\c R_T=\O(T)$, it is important to note that when $-\c R_T=\O(T^{1/2})$, i.e., when LLP does not outperform the benchmark by more than $T^{1/2}$, then for perfect predictions we achieve $\c R_T=\O(1)$, $\c V_T=\O(T^{1/4})$ (setting $\beta=0$), and for worst-case predictions it is $\c R_T, \c V_T=\O(T^{1/2})$. On the other hand, if LLP does outperform the benchmark consistently ($\forall \,T$) by at least $\Omega(T^{1/2})$, we achieve negative regret and $\c V_T\!=\!O(T^{2/3})$. The general case, for which the above two Corollaries hold, is when the sample path is such that LLP bounces above and below the performance of $\{x^\star\}$. A schematic description of these cases is included in the Appendix and the different achieved rates by LLP are summarized in Table \ref{table-bounds}.

Concluding, it is worth discussing the inherent difficulties of the problem at hand. Namely, one might argue that we could directly apply the optimistic FTRL result of \cite{mohri-aistats2016} (or our equivalent Lemma \ref{predictionsb}) both to the primal and to the dual update, and combine the results to bound $\c R_T$ and $\c V_T$. The interested reader, however, can verify that this straightforward strategy leads to much worse bounds. Our approach instead is to apply the optimistic FTRL only to the dual update and carefully reconstruct tighter bounds for the Lagrangian, while using a fixed-point iteration to find the exact (minimum) growth rate of $\|\ll_T\|$. Moreover, unlike prior works such as \cite{victor} or \cite{jennaton}, we update $\{x_t\}$ using $\{\lambda_t\}$ (instead of $\lambda_{t-1}$); and then update $\{\lambda_{t+1}\}$ using the newly calculated $\{x_t\}$. This strategy facilitates the inclusion of predictions as we only need to predict $x_t$ and not $\lambda_t$. Besides, it is exactly this circular relation between the primal and dual variables which renders the inclusion of predictions in OCO with budget constraints fundamentally different from the respective OCO problem without time-varying constraints. 
 
\subsection{Proof of Theorem \ref{Theorem2}} \label{sec:Th2proof}

Our strategy is to bound the growth rate of the dual vector norm and use it to bound $\c R_T$ and $\c V_T$. First, we define its minimum growth rate $k=\min\{\varphi :\|\ll_t\|\!=\!\O(t^\varphi\}$, and introduce $\LL_t=\max\left\{ \|\ll_i\|: i\leq t \right\}$, where $\LL_t\!\leq\! \LL_T\!=\!\O(T^k)$, $\forall t\leq T$. Using the closed-form solution\footnote{Eq. \eqref{dual-update} simplifies to $\lambda_{t+1}=\arg\min_{\ll\in\mathbb R_+^d }\big\{ \frac{\|\lambda\|^2}{2a_t}-\ll^\top v\big\}$ with $v=\p g_{t+1}(\p x_{t+1})+\sum_{i=1}^t g_i(z_i)$.} of \eqref{dual-update}, we can write: 
\begin{align}
\!\!\!\!\|\ll_{t+1}\|\!= \!\bigg\|\! \bigg[ a_{t}\Big(\sum_{i=1}^{t}\! g_i(z_i)\!+\!\p g_{t+1}(\p x_{t+1})\Big)\! \bigg]_+ \bigg\| \!\leq\! a_{t}\big(V_{t}^z \!+\! G\big) \!\!\!  \label{lambda-rate2}
\end{align}
where we used that $\|\p g_t(z)\|, \| g_t(z)\|\!\leq\! G$, $\forall t$, and the triangle inequality.  Also, we can write:
\begin{align}
\!\!\!\!\!h_{1:T}\!=\!\sum_{t=1}^T\! \left\| \epsilon_t\!+\!\ll_t^\top\delta_t\right\|\! \leq\! T(E_m\!\!+\! \Delta_m \Lambda_T)\!=\! \O(T^{k+1})\!\! \label{bound-eq1}
\end{align}
We will use this bound and the definition of $\ll_t$, which ties it to $V_t^z$ to find a smaller growth rate than the one we would get by directly bounding $V_t^z$ in \eqref{lambda-rate2}. Indeed, starting from \eqref{lambda-rate2} and using \eqref{vtz-bound}, we have:
\begin{align}
\|\ll_{T+1}\| &\!\leq \! a_{T}\sqrt{ 2\big(B_T-R_T\big)/a_{T-1}}+ a_TG\,\, \stackrel{a_{T-1}\geq a_T}\Longrightarrow \notag \\ 
\|\ll_{T+1}\| &\!\leq \!\sqrt{ 2a_T\big(B_T-R_T\big)}+ a_TG. \label{lambda-rel}
\end{align}
Next, note that based on the definition of $a_t$ and the fact that we consider worst-case predictions (which increase linearly with $T$), the following inequalities hold concurrently:
\begin{align}
&a_{T}\leq \frac{a}{ \sqrt{4G^2+\SumN \xi_t^2}}=\O(T^{-\frac{1}{2}})\notag \\
&a_{T} \leq \frac{a}{T^\beta}=\O(T^{-\beta}). \notag
\end{align} 
Thus, it follows that:
\begin{align}
a_T=\O\big(T^\theta\big),\,\,\quad \theta=\min\left\{- \beta, -\frac{1}{2} \right\}\leq 0. \label{bound-eq3}
\end{align}
Similarly, the following inequalities hold:
\begin{align}
&\SumN a_{t-1}\xi_t^2\leq \SumN \frac{a\xi_t^2}{\sqrt{\SumN \xi_i^2}}\leq 2a \sqrt{ \SumN \xi_t^2}=\O(T^{\frac{1}{2}}) \notag \\
&\SumN a_{t-1}\xi_t^2\leq \SumN a\frac{\xi_t^2}{t^\beta}\leq\frac{4a G^2}{1-\beta}T^{1-\beta} =\O(T^{1-\beta}), \notag 
\end{align}
where we used \cite[Lemma 3.5]{auer2002}, the identity $\SumN t^{-\beta}\!\leq\! T^{1-\beta}/(1\!-\!\beta)$ (Lemma \ref{SumNointegral} in Appendix) and $\xi_{t}= 2G, \forall t$. Hence:
\begin{align}
\SumN a_{t-1}\xi_t^2 = \O(T^n),\quad n=\min\Big\{ \frac{1}{2},\,1-\beta\Big\}\leq 1. \label{bound-eq2}
\end{align}

Finally, replacing $B_T$ in \eqref{lambda-rel} with its definition from Theorem \ref{Theorem1}, we arrive at:
\begin{align}
\|\ll_T\|=\O\left( \max\left\{ T^{\frac{k+1+2\theta}{4}},\,\,\, T^{\frac{n+\theta}{2}}, \,\,\,T^{\frac{1+\theta}{2}}\,\,\,T^{\theta} \right\} \right),
\end{align}
where we used the worst case bound $-\c R_T=\!\O(T)$ and $a_TG=\O(T^\theta)$. To find the dominant term, note that, since $\beta\in[0,1)$, it is $n\leq 1$, and hence $(1+\theta)/2 \! \geq\! (n+\theta)/2$, thus we omit the second term. Also, $\theta\leq 0$ and hence we can omit the last term; and finally, from \eqref{lambda-rate2} we observe that $k\!\leq 1+\theta \!\leq \! 1$, thus, the third term is larger than the first, and we conclude with $\|\ll_T\|\!=\!\O(T^k)\!=\!\O(T^{(1+\theta)/2})$. 

Having found the growth rate of $\|\ll_T\|$ to be $k=(1+\theta)/2$, we use \eqref{bound-eq1} and \eqref{bound-eq2}  to refine the bounds:
\begin{align}
\!\!\!\!h_{1:T}\!=\!\O\left(T^{\frac{3+\theta}{2}}\right),\, B_T\!=\!\O\left(\! \max\left\{ T^{\frac{3+\theta}{4}}, T^n \right\} \!\right) \!\!\label{eq4}
\end{align}
and these conclude the proof of  Theorem \ref{Theorem2}\textbf{(a)}.  For the constraint violation $\c V_T$, observe first that $(1/a_T)=$
\begin{align}
\frac{ \max\!\Big\{\!\! \sqrt{ G^2 + \SumN \xi_t^2 }, T^\beta \! \Big \}}{a}=\O(T^{\nu}),\,\, \nu\!=\max\left\{\beta, \frac{1}{2}\right\} \notag  
\end{align} 
and hence holds $\nu=-\theta$. Using this bound along with \eqref{eq4} and $-\c R_T\!=\!\O(T)$, we get from Theorem \ref{Theorem1}(\textbf{b}):
\begin{align*}
&\c V_T =\O \left( \max\left\{ T^{\frac{3(1-\theta)}{8}}, T^{\frac{n-\theta}{2}}, T^{\frac{1-\theta}{2}}, T^{\frac{3+\theta}{4}} \right\}\right).
\end{align*}
We conclude by noticing that:
\[
\frac{1-\theta}{2}> \frac{3(1-\theta)}{8} \quad \text{and} \quad \frac{1-\theta}{2}\geq \frac{n-\theta}{2},
\]
and observing that conditioning on the value of $\beta$, we get the bounds in Theorem \ref{Theorem2}.

\section{Less Computations And Predictions} \label{sec:less}

We discuss next how to reduce the computation and memory requirements of  LLP by using non-proximal primal regularizers; and the impact of  linearizing the constraint functions on the required predictions.

\subsection{LLP with Non-proximal Regularizers}

In this new algorithm, called LLP2, we use the same dual regularizer \eqref{dual-reg} and update \eqref{dual-update}, but the general primal regularizer:
\begin{align}
&r_0(x)=\mathbf{I}_{\X}(x)\,\,\text{and}\,\, r_t(x)=\sigma_t\|x\|^2/2,\,\, \forall t\geq 1,\,\text{with:}\notag \\	
&\sigma_t=\sigma\left( \sqrt{h_{1:t} +\mu_{t+1}} - \sqrt{h_{1:t-1}+\mu_{t}} \right)  \label{primal-reg2}
\end{align}
where $\mu_{t}\!\triangleq\! E_{m}\!+\!a_{t-1}Gt\DD_{m}$. The new updates are:
\begin{align}
	&\!\!x_t\! =\! \arg \min_{x\in \RR^N} \ts  \left\{\sum_{i=0}^{t-1} \L_i(x,\ll_i) \!+\p c_t^\top x \!+\ll_t^\top \p g_t(x) 	\right\}, \!\label{primal-update2} \\
	&\!\!z_t \!=\! \arg \min_{x\in \RR^N} \ts \left\{ \sum_{i=0}^{t-1} \L_i(x,\ll_i) \!+c_t^\top x \!+\ll_t^\top g_t(x)\right\},\!\! \label{prescient-update2}
\end{align}
where $\L_t(x,\ll)$ is defined using $r_t(x)$ from \eqref{primal-reg2}, and note the off-by-one regularizer of \eqref{prescient-update2} compared to \eqref{prescient-update}. 

These non-proximal regularizers facilitate solving for $\{x_t\}$ and $\{z_t\}$ since $r_{1:t}(x)\!=\!\sigma_{1:t}\|x\|^2/2$ involves only one quadratic term and can be represented in constant memory space, unlike $r_{1:t}(x)\!=\!\sum_{i=1}^t \sigma_i\|x\!-x_i\|^2/2$ that expands with time. On the other hand, non-proximal updates yield looser bounds, cf. \cite{mcmahan-survey17}, and require a new saddle-point analysis. Interestingly, they do not worsen the growth of LLP2, but do prevent it from achieving $\c R_T=\!\O(1)$ for perfect predictions. 
\begin{mdframed}
\begin{theorem}\label{Theorem3}
Under Assumptions (\textbf{A1})-(\textbf{A4}) and with $\{r_t\}$ and $\{q_t\}$ satisfying \eqref{primal-reg2} and \eqref{dual-reg}, LLP2 ensures for every $x^\star\!\in\! \X_T$: 
\begin{align} 
\!\!\!\!&\c R_T\!\le\! \widehat B_T\!\triangleq\! 2\Big(\!\sigma D^2 \!+\!\frac{L_f}{\sigma}\!\Big)\!\sqrt{h_{1:T}\!+\!\mu_{T+1}} \!+\!\sum_{t=1}^T \frac{\xi_t^2}{\phi_{0:t-1}}   \notag\!\!\! \\  
\!\!\!\!&\c V_T\!\le \! \sqrt{2\phi_{0:T-1}(\widehat B_T\!-\!\c R_T)}\!+\!\frac{2L_g}{\sigma}\!\sqrt{h_{1:T}\!+\!\mu_{T+1}} \notag\!\!\! 
\end{align}
where $\mu_{T}=E_m+a_{T-1}GT\Delta_m$.
\end{theorem}
\end{mdframed}
The proof of Theorem \ref{Theorem3} can be found in the Appendix. 

\begin{corollary}\label{llp2-cor1}
LLP2 achieves the same regret bounds as those described in Theorem \ref{Theorem2} for LLP in the general case, and under perfect predictions it ensures:
\begin{align*}
	\c R_T\!=\!\O\Big(T^{\frac{1-\beta}{2}}\Big),\,\,\,\,\c V_T\!=\O\Big(T^{\frac{1+\beta}{2}}\Big),\,\,\,\,\,\,\beta\!\in\![0,1).
\end{align*}
\end{corollary}
For example, LLP2 with perfect predictions and $\beta=0$ achieves $\c R_T=\O(T^{1/2})$ and $\c V_T=\O(T^{1/2})$; and with $\beta=1/3$ yields $\c R_T=\O(T^{1/3})$ and $\c V_T=\O(T^{2/3})$. Hence, it outperforms the state-of-the-art constrained-OCO algorithms with no predictions, but does not perform as well as LLP for perfect predictions.

\subsection{LLP with Linearized Constraints} 

Another way to reduce the computation load of LLP is to linearize the constraint function. This, however, is not trivial since we cannot recover $\c V_T$ by simply using the convexity of $\{g_t\}$, as we do with $\c R_T$ and the linearization of $\{f_t\}$. Hence, we use linear proxies for the constraint and its prediction in each round $t$:
\begin{align}
	&g_t^{\ell}(x)=g_t(x_t)+ \grad g_t(x_t)^\top (x-x_t), \label{lin1} \\  
	&\p g_t^{\ell}(x)=\p g_t(\p x_t)+ \grad \p g_t(\p x_t)^\top (x-\p x_t). \label{lin2} 
\end{align}
LLP with linearized constraints runs similarly to Algorithm \ref{alg:llp}, but uses predictions $\p c_t$, $\grad \p g_t(\p x_t)$ and $\p g_{t+1}(\p x_{t+1})$, i.e., does not need to predict the entire constraint function --- nor $\p x_t$, despite appearing in \eqref{lin2}. Interestingly, this does not affect its performance.
\begin{lemma}\label{lem:linearized}
LLP with linearized constraints and predictions given by \eqref{lin1}-\eqref{lin2}, achieves the $\c R_T$, $\c V_T$ bounds and convergence rates in Theorems \ref{Theorem1} and \ref{Theorem2}, respectively.
\end{lemma}
The proof of the Lemma and the details of the linearized LLP can be found in the Appendix. Whether it is more challenging to predict the next constraint gradient or the next constraint function, is a question pertaining to the problem at hand, and one can select the version of LLP that is more suitable for that.

\section{Linearly-Perturbed Constraints}\label{sec:pert}
In this section we consider the special type of constraints that are linearly-perturbed, which was studied first in \cite{victor}. In detail, in this case the constraints and their predictions are:
\begin{align}
g_t(x)=g(x)+b_t \quad \text{and} \quad \p g_t(x)=g(x)+\p b_t. \label{constr-linear}
\end{align}
where $b_t, \p b_t \in \mathbb R^d$ are the unknown per-slot perturbations that are added to the fixed (and known) function component $g(x)$. This simplification has important ramifications for the analysis and, eventually, improves the bounds of LLP as follows:
\begin{lemma}\label{lem:linear-const} Under the conditions of Theorem \ref{Theorem1}, with constraints and predictions given by \eqref{constr-linear}, LLP ensures:
\begin{align*}
&\c R_T\leq 0,\,\,\, \c V_T =\c O(T^{(1+\beta)/2}), \,\,\,\,\,\,\text{perfect predict.}, \beta\!\in\! [0,1) \\
&\c R_T=\c O(\sqrt T ),\,\,\, \c V_T=\c O(T^{5/8}),\,\,\text{worst predictions}, \beta\!=\!1/2
\end{align*}
\end{lemma}

This result improves the bounds for the case of general constraints of the previous section, and yields only $1/4$ worse constraint violation than the bounds in \cite{victor} -- which however cannot benefit from predictions. On the contrary, we see that with LLP, due to its prediction-adaptive steps, we get no regret for perfect predictions and, interestingly, the constant factors of the bounds shrink commensurately with the predictions' accuracy. This becomes clear if we express the bounds as:
\begin{align}
&\!\!\c R_T\!\leq B_T\!\triangleq \!A_1\sqrt{h_{1:T}}+\min\left\{\! 2a\sqrt{\sum_{t=1}^T\xi_t^2 },\, A_2T^{1-\beta}	\right\} \label{regret-linear-detail} \\
&\c V_T\!\leq \left[ A_3\max\left\{ K_T, T^\beta	\right\}\big(B_T-R_T\big)  \right]^{\frac{1}{2}} +A_4 \sqrt{h_{1:T}}  \label{regret-constr-detail} 
\end{align}
where we have defined the parameters:
\begin{align*}
&A_1\!=2\sigma D^2\!+\!(2L_f/\sigma) , A_2\!=4aG^2/(1\!-\!\beta), A_3=2/a  \notag\\
&A_4=2L_g/\sigma, h_{1:T}=\sum_{t=1}^T \|\epsilon_t\|, K_T=\Big[ G^2 + \sum_{t=1}^T \xi_t^2\Big]^{1/2}
\end{align*}
and note that in this case the quantity $h_{1:T}$ does not depend on the dual vectors. This is due to the fact that the perturbations do not affect the primal step (see details Sec. \ref{appendix:linear-const} of the Appendix), which disentangles -- to some extent -- the primal and dual iterations. The proof of the lemma and the details for deriving \eqref{regret-linear-detail} and \eqref{regret-constr-detail} can be found in the Appendix.

\section{Conclusions}\label{sec:conclusions}
LLP differs from related algorithms since the primal and dual updates are \textit{lazy}. This  allows an FTRL-based analysis, which is widely used in fixed-constraints OCO algorithms ($x\in \X$) but is new in the context of time-varying constrained OCO ($x \in \X_T$). The LLP order bounds, even with worst-case or no predictions, are competitive with existing algorithms while dropping several restrictive and often impractical assumptions these are using. Indeed, prior algorithms with time-varying constraints require strongly convex cost and constraint functions or linearly-perturbed fixed constraints, and rely on the Slater condition. Other proposals assume {time-invariant constraints}, still rely on the Slater condition and on a fixed and known time horizon $T$ (which most often is not available); and need access to all Lipschitz constants and constraint bounds. Crucially, these prior works do not benefit from the availability of (potentially inaccurate) predictions. When the latter are perfect, LLP achieves $\c R_T=\O(1)$ and $\c V_T=\O(\sqrt T)$. Last but not least, our framework is unified as it and can run without predictions (setting them zero) since we impose no assumptions on their quality, and can be applied to problems with time-invariant constraints.

\bibliographystyle{IEEEtran}
\bibliography{references-journal}

\section{Appendix} \label{sec:appendix} 

The Appendix includes the missing proofs from the main document; the supporting lemmas and their proofs; and additional discussion for the main results. 

\begin{figure*}[t!]
	\centering
	{\includegraphics[width=5.86in]{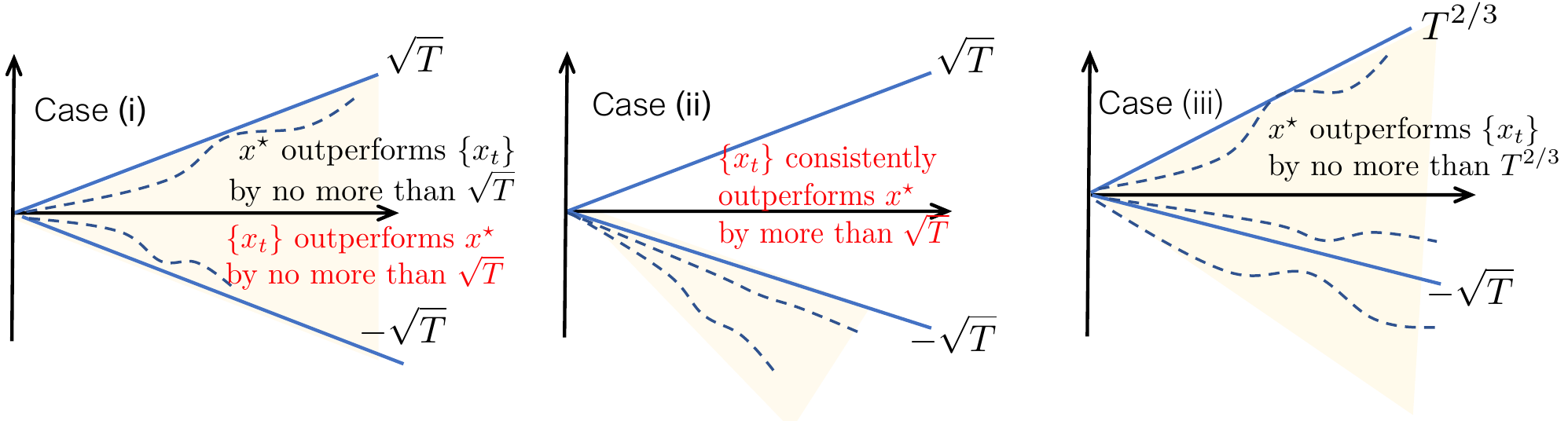}}		
	\caption{Performance region and cases for the Regret of LLP, under worst-case (or, no available) predictions.}
	\label{Fig:cases}
\end{figure*} 

\subsection{Performance Cases of LLP}

We start by discussing the different cases regarding the performance of LLP in order to facilitate the reader understanding the consequences of Theorems 1 and 2. Figure \ref{Fig:cases} summarizes the three possible scenarios. Case (i) is realized when $-\c R_T=\O(T^{1/2})$, i.e., when LLP does not outperform $x^\star$ by more than this growth rate, and here LLP achieves quite compelling rates, $\c R_T, \c V_T=\O(T^{1/2})$ and zero regret for perfect predictions. Case (ii) arises when the condition $-\c R_T=\O(T^{1/2})$ is consistently violated and in fact yields even better performance in terms of regret, while maintaining the general $\c V_T=\O(T^{2/3})$. And finally, Case (iii) arises when the above condition might be violated during some time intervals and sample paths, but not consistently; and for this scenario the general bounds $\c R_T, \c V_T=\O(T^{2/3})$ hold. In all cases, the constant factors of the regret diminish as the predictions' quality improves.

\subsection{Proof of Lemma \ref{lem:impossibility}} 
\begin{mdframed}
\textbf{Lemma \ref{lem:impossibility}}. No online algorithm can achieve concurrently sublinear regret and constraint violation, $\c R_T=o(T)$, $\c V_T=o(T)$, w.r.t. $x^\star \in X_T^{\text{max}}$, even if the algorithm selects $x_t$ with knowledge of $f_t, g_t$.
\end{mdframed}
\begin{proof}
We provide an opponent strategy that ensures there is an increasing sequence $t(1),t(2),\ldots$ of rounds with either $\c R_{t(i)} \ge t(i)/8$ or $\c V_{t(i)} \ge t(i)/8$. Our opponent will select $f_{t+1},g_{t+1}$ based only on $x_1,\ldots, x_t$. Hence the impossibility result holds even if the player knows $f_{t+1},g_{t+1}$ on round $t+1$.  

Consider the domain $\X=[0,1]$. The cost functions are linear and the pair $(f_t,g_t)$ is always one of $p\!=\!(-x,-1)$ or $q\!=(-2x,2x-1)$. Before giving the opponent strategy we make some general observations. To derive the set $\SumN g_i(x)\!\le 0$ suppose the opponent plays $p$ exactly $n$ times and $q$ exactly $T\!-n$ times. Then we have:
\begin{align*}
\SumN g_t(x)= -n + (T-n)(2x-1)  = 2(T-n)x -T
\end{align*}
Hence the constraint set $G_T = \big\{x \in \X: \SumN g_i(x) \le 0\big \}$ is the part of $[0,1]$ with 
$$x \le \frac{T}{2(T-n)} = \frac{1}{2} + \frac{n}{2(T-n)}.$$

In particular for $n \ge T/2$ the second term is at least $1/2$ and $G_T=[0,1]$. Since $f_t$ are negative linear, the regret is with respect to $x^\star\!=\!1$. For $f_t(x) \!=\! -x$ we have $f_t(x_t)\!-f_t(x^\star) \!= 1-x_t$ and for $f_t(x) = -2x$ we have $f_t(x_t)-f_t(x^\star) = 1-2x_t$. In each case  $f_t(x_t)-f_t(x^\star) \ge 1-x_t$ and so $\c R_T \ge  \SumN (1-x_t)$. 

Now suppose the rounds are broken into blocks $\{1,2,\ldots\} = I_1 \cup J_1 \cup I_2 \cup J_2 \cup \ldots$ with each $|I_n|=|J_n|$. Define each $I = \bigcup_n I_n$ and $J = \bigcup _n J_n$ and $\ol x_{t-1} = \frac{1}{t-1} \sum_{i=1}^{t-1} x_i$. Assume the opponent has the strategy:
\begin{enumerate}[leftmargin=5mm]
	\item Each $(f_t,g_t)= q \iff t\in I$.
	\item Each $(f_t,g_t)= p  \iff t \in J$.
	\item For each $n \in I$ we have $\ol x_{t-1} \ge 3/4$.
	\item For each $m$ and each $t\!=\!\min J_m$ we have $\ol x_{t-1} \!<\! 3/4$. 
\end{enumerate}$ $ 
There are two cases to consider. First assume (a) there are infinitely many $I_m,J_m$. Since each $|I_m|=|J_m|$ we see on the final turn $t(m)$ of each $J_m$ the opponent has played $p$ exactly $n = T/2$ times. Hence the above says $G_{t(m)} \!= [0,1]$ and $x^\star \!= 1$ and the regret is $R_{t(m)}\! \ge\! \sum_{i=1}^{t(m)} (1-x_i)$. On turn $s(m)\! = \max I_m$ the regret is at most:
\begin{align}  
&\sum_{i=1}^{s(m)} (1-x_i)= s(m) -\sum_{i=1}^{s(m)} x_i = s(m) -s(m)  \ol x_{s(m)}\notag \\
&= s(m)(1- \ol x_{s(m)}) \ge \frac{s(m)}{4} \ge \frac{t(m)}{8}. \notag
\end{align}
where the first inequality uses (4) and the second uses  $t(m) =s (m) + 1 \ge t(m)/2$. Since there are infinitely many $t(1),t(2),\ldots$ there are infinitely many turns $t$ with $R_t \ge t/8$. Hence the regret is $\Omega(T)$. 

Now assume (b) there are only finitely many $I_m,J_m$. Since each $|I_m| = |J_m|$ the blocks are $I_1,J_1,\ldots, J_{m-1},I_m$ with $|I_m|=\infty$. To see the violation is $\Omega(T)$ observe (3) gives $\ol x_s \ge 3/4$ for $s = \max J_{m-1}$. By (1) the opponent plays $q= (2x,-2x+1)$ on turns $s+1,s+2,\ldots $  Thus for $T \ge 4s$ the constraint violation is:
\begin{align}
\SumN g_t(x_t) &= \sum_{t=1}^s g_t(x_t) +   \sum_{t>s}^{T} g_t(x_t) \ge -s + \sum_{t>s}^{T} (2x_t-1) \notag\\
&= - T  + 2 \sum_{t>s}^{T} x_t = - T -  \sum_{t=1}^{s} x_t  + 2 \sum_{t=1}^{T} x_t  \notag \\
&\ge -(T+s) + 2 \ol x_T  \ge -(T+s) + \frac{3}{2}  T \notag
\end{align}
where the last inequality uses (3). Since $T \ge 4s$ the RHS is at most $T/2 -s \ge T/2 - T/4 = T/8$. Hence we can take $t(i) = 4s +i$.

To complete the proof we give an opponent strategy that satisfies the four conditions. \cite{tsitsiklis} suggest the following. 
\begin{enumerate}
	\item[(i)] Play $q$ on turns  $ \min I_n, \min I_n+1, \ldots, m$ where $m$ is the first turn with $\ol x_m < 3/4$. 
	\item[(ii)]End $I_n$ and begin $J_n$.
	\item[(iii)] Play $p$ over the next $|I_n|$ turns. 
	\item[(iv)]End $J_n$ and begin $I_{n+1}$.
\end{enumerate} 
Note the decision on turn $m+1$ to end $I_n$ depends only on the average $\ol x_m$ of $x_1,\ldots, x_m$. Hence the opponent does not need to see the player's current move to implement the strategy. Equivalently the player is allowed to see the opponent's next move. 
\end{proof}

\subsection{Proof of Lemma \ref{2ssclaim2b}}

Using the property of the proximal regularizers, $x_t=\arg\min_{x} r_t(x), \forall t$, we can expand \eqref{primal-update} and write:
\begin{align}
x_t =&\arg\min_{x\in \RR^N}  \bigg\{r_{0:t-1}(x) \!+c_{1:t-1}^\top x +  \sum_{i=1}^{t-1} \ll_i^\top g_i(x)\notag \\ 
&+ \p c_t^\top x \!+\ll_t^\top \p g_t(x)\!\bigg\} \stackrel{\text{add}\,\,\,r_t(x)}\Longrightarrow \notag \\
x_t =&\arg\min_{x\in \RR^N}  \bigg\{\!r_{0:t}(x) +c_{1:t-1}^\top x \!+  \sum_{i=1}^{t-1} \ll_i^\top g_i(x) \notag \\
&+ \p c_t^\top x +\ll_t^\top \p g_t(x)\bigg\}, \notag
\end{align}
where adding the $t$-round regularizer $r_t(x)$ does not change the minimizer of the RHS argument -- and it is easy to see this using a contradiction argument. 

Now, recall that the prescient action is:
\begin{align}
z_t &=\arg\min_{x\in \RR^N}  \left\{r_{0:t}(x) +c_{1:t}^\top x +  \sum_{i=1}^{t} \ll_i^\top g_i(x) \right\}. \notag
\end{align}
Applying Lemma 7 from \cite{mcmahan-survey17}, with:
\begin{align}
&\phi_1(x)=r_{0:t}(x) +c_{1:t-1}^\top x +  \sum_{i=1}^{t-1} \ll_i^\top g_i(x) + \p c_t^\top x +\ll_t^\top \p g_t(x) \notag\\
&\phi_2(x)=r_{0:t}(x) +c_{1:t}^\top x +  \sum_{i=1}^{t} \ll_i^\top g_i(x) \quad \text{and} \notag \\
&\psi(x)=(c_{t}-\p c_t)^\top x +  \ll_t^\top\big(g_t(x)-\p g_t(x)\big) \notag
\end{align}
and recalling that the regularizer $r_{0:t}(x)$ is 1-strongly-convex w.r.t. norm $\|x\|_{(t)}\!=\!\sqrt{\sigma_{1:t}}\|x\|$ that has  dual norm $\|x\|_{(t),\star}\!=\!\|x\|/\sqrt{\sigma_{1:t}}$, we can write:
\begin{align*}
&\|x_t-z_t\|_{(t)}\leq \| c_t - \p c_t + \ll_t^\top \big( \grad g_t(x_t) - \grad \p g_t(x_t) \big)\|_{(t),\star} \Rightarrow\\
&\sqrt{\sigma_{1:t}}\|x_t-z_t\|\leq \frac{\| c_t - \p c_t + \ll_t^\top \big( \grad g_t(x_t) - \grad \p g_t(x_t) \big)\|}{\sqrt{\sigma_{1:t}}} \Rightarrow\\
&\|x_t-z_t\|\leq =\frac{\|\e_t+\ll_t^\top \dd_t\|}{\sigma_{1:t}}
\end{align*}

\subsection{Proof of Lemma \ref{predictionsb} (Optimistic FTRL)}
In the proof of Theorem \ref{Theorem1} we relied on \cite[Theorem 2]{mohri-aistats2016}. However, that work considers a learning problem over a compact convex set, while the dual update to which we apply this result has an unbounded decision space $\lambda\in \mathbb R_+^d$. This indeed does not pose a problem for our analysis. Firstly, one can see that for the standard FTRL analysis it suffices to have closed sets, see for example \cite{mcmahan-survey17}. Secondly, the boundedness of the set is useful when we need to upper-bound the term $q_{0:T-1}(\ll)$ that appears in the RHS of \eqref{eq:dual1}. In our analysis, this is not necessary as we cancel this term by setting $\lambda=0$.

To complete this discussion, we provide here an alternative proof for \cite[Theorem 2]{mohri-aistats2016} that makes clear it is valid even if the decision set is not bounded. We note that this is presented in terms of the primal variables and functions $\{f_t(x_t)\}$ to streamline the presentation, but the application to the dual variables and dual updates is straightforward.

\begin{mdframed}
\begin{lemma}\label{predictionsb}
Let $\{r_t\}$ be a sequence of non-negative regularizers, and let $\p c_t$ be the learner's estimate of $c_t=\grad f_t(x_t)$. Assume also that the function $\kappa_{0:t-1}:x\mapsto c_{1:t-1}^\top x + \p c_{t}x + r_{0:t-1}(x)$ is 1-strongly convex w.r.t.  norm $\|\cdot\|_{(t-1)}$, and consider the update $x_t=\arg\min_{x\in \X} \kappa_{0:t-1}(x)$, where $r_0(x)=\mathbf I_{\X}(x)$ and $\p c_1=0$. Then the following regret bound holds:
\begin{align}
\c R_T\leq r_{0:T-1}(x) \!+ \SumN \|c_t\!-\p c_t\|_{(t-1),\star}^2, \forall x\in \X\notag
\end{align}
\end{lemma}
\end{mdframed}
\begin{proof}
Using the auxiliary functions $m_t(x)=r_{t-1}(x)+\p c_t^\top x$ and $n_t(x)=c_t^\top x-\p c_t^\top x$, we can write $x_{t}=\arg\min_{x\in\X}\{m_{1:t}(x)+n_{1:t-1}(x)\}$ and $z_{t}=\arg\min_{x\in\X} \{m_{1:t}(x)+n_{1:t}(x)\}$ which correspond to the primal and prescient updates for the non-proximal version of the optimistic FTRL. We first prove the following relation using induction:
\begin{align}
\SumN \! m_t(x_t)+n_t(z_t)\!\leq \! m_{1:T}(x^\star)\!+n_{1:T}(x^\star), \,\, \forall x^\star \in \mathcal X.  \label{btl}
\end{align}
For $t=1$,  it is:
\begin{align*}
&m_1(x_1)+n_1(z_1)=r_0(x_1)+\p c_1^\top x_1 +c_1^\top z_1- \p c_1^\top z_1 \notag \\
&\leq r_0(x^\star) +\p c_1^\top x^\star + c_1^\top x^\star- \p c_1^\top x^\star = m_1(x^\star)+n_1(x^\star)
\end{align*}
which holds since $r_0(x_1)=r_0(x^\star)=0$, $\p c_1^\top x_1=0$, and $z_1=\arg\min_{x\in\X}\big\{m_1(x)+n_1(x)\big\}$. Assume it holds for $t=\tau$ and add to both sides $m_{\tau+1}(x^\star)\!+\! n_{\tau+1}(x^\prime)$ for some $x^\star, x^\prime \in \mathcal X$:
\begin{align*}
	&\sum_{t=1}^\tau m_t(x_t)\!+m_{\tau+1}(x^\star)+\sum_{t=1}^\tau n_t(z_t)\!+n_{\tau+1}(x^\prime)\notag \\
	& \leq m_{1:\tau+1}(x^\star) +n_{1:\tau}(x^\star)+n_{\tau+1}(x^\prime)\\
	&\text{set}\,\,x^\star\!=\!x_{t+1},\,\,\, x^\prime\!=\!z_{t+1},\,\,\text{to get:} \\
	&\sum_{t=1}^{\tau+1}\!m_t(x_t)\!+\! n_t(z_t)\!\leq\!m_{1:\tau+1}(x_{t+1}) \!+\!n_{1:\tau}(x_{t+1})\!+\!n_{\tau+1}(z_{t+1}) \\
	&\leq m_{1:\tau+1}(z_{t+1})+n_{1:\tau+1}(z_{t+1})\,\,\,\,\,\,(\text{by definition of} \,\, x_{t+1})\\
	&\leq  m_{1:\tau+1}(x^\star)+n_{1:\tau+1}(x^\star)\,\,\,\,\,\,\,\,\,\,\,\,\,\,\,\,(\text{by definition of} \,\, z_{t+1})
\end{align*}
which concludes our induction step. Hence, we proved \eqref{btl}. 

Replacing $m_t(x)$ and $n_t(x)$, dropping the non-negative term $\SumN r_t(x_t)$ in the LHS, adding $\SumN c_t^\top x_t$ to both sides and rearranging, we eventually get:
\begin{align}
	\c R_T &\leq r_{0:T}(x^\star) + \SumN \Big( c_t^\top(x_t-z_t)\Big)+ \SumN \Big(\p c_t^\top (z_t-x_t) \Big) \notag \\
	&\leq r_{0:T}(x^\star) + \SumN \big(c_t-\p c_t\big)^\top(x_t-z_t) \\
	&\leq r_{0:T}(x^\star) + \SumN \big\|c_t-\p c_t\|_{(t-1),\star}\|x_t-z_t\|_{(t-1)} \label{last-eq}
\end{align}
where in the last step we used the Cauchy-Schwarz inequality. For the term $\|x_t-z_t\|_{(t-1)}$ we can apply \cite[Lemma 7]{mcmahan-survey17} with:
\begin{align*}
&\phi_1(x)=r_{0:t-1}(x) +c_{1:t-1}^\top x + \p c_t^\top x \\
&\phi_2(x)=r_{0:t-1}(x) +c_{1:t-1}^\top x + \p c_t^\top x  + ( c_t^\top x - \p c_t^\top x ) \\
&\psi(x)=c_t^\top x - \p c_t^\top x
\end{align*}
to get the bound $\|x_t-z_t\|_{(t-1)}\leq \|c_t - \p c_t\|_{(t-1), \star}$. Replacing in \eqref{last-eq} we conclude the proof noting that we did not use boundedness of $\X$ in any step of the proof.

\end{proof}

\subsection{Proof of Lemma \ref{SumNointegral}}

This Lemma is also used in \cite{jennaton}; we include its proof here for completeness.

\begin{mdframed} 
\begin{lemma}\label{SumNointegral}
For  $d \in (0,1)$ we have $$\SumN t^{-d} \le \frac{T^{1-d}}{1-d}$$  
\end{lemma}
 \end{mdframed}

\begin{proof}
Let $F(x) = \lfloor x+1\rfloor^{-d}$  be defined over $[0,\infty)$. Clearly $F(x) = (n+1)^{-c}$  for each $x \in [n,n+1)$ and $n =0,1,2,\ldots$, and so the sum on the left equals $\int_0^{T} F(x) dx$. Since $\lfloor x+1\rfloor \ge x$ and $d >0$ we have $F(x) = \lfloor x+1\rfloor^{-d} \le x^{-d}$ and $\int_0^T F(x) dx \le \int_0^T x^{-d} dx = \frac{T^{1-d}}{1-d}$.
\end{proof}

\subsection{Proof of Theorem \ref{Theorem3}}

First, note we can directly obtain the bound:
	\begin{align*}
		\|\ll_{T}\| &\leq \left\| \left[ a_{T-1}\left(\sum_{i=1}^{T-1} g_i(z_i)+\p g_{T}(\p z_{T})\right) \right]_+\right\|\leq a_{T-1}TG 
	\end{align*}
	where the RHS term appears in $\mu_{T}$. Hence, the strong convexity of LLP2 can be lower-bounded as:
	\begin{align}
		\sigma_{1:t}&=\sigma \sqrt{ h_{1:t}+\mu_{t+1}} \geq \sigma\sqrt{ h_{1:t+1}}. \label{eq5}
	\end{align}
	Next, we replace Lemma \ref{2ssclaim2b} with an updated bound.
	\begin{lemma}
		For the actions $x_t$ and $z_t$ obtained by \eqref{primal-update2} and \eqref{prescient-update2}, respectively, it holds:
		\begin{align}
			\!\!\|x_t\!-z_t\|\leq \frac{ \|\epsilon_t\!+\ll_t^\top\dd_t\| }{\sigma_{1:t-1}} \stackrel{\eqref{eq5}} \leq \frac{ \|\epsilon_t\!+\ll_t^\top\dd_t\| }{  \sqrt{\sum_{i=1}^t \|\epsilon_i\!+\ll_i^\top\dd_i\|}}
		\end{align}
	\end{lemma}
	
	It is easy to see that since the dual update has not changed, equation \eqref{eq:dual1} holds as is, and we can readily obtain \eqref{eq1}, and  eventually write:
	\begin{align}
		\c R_T&\leq 2D^2 \sigma_{1:T} \!+ \frac{2L_f}{\sigma}\sqrt{h_{1:T}}+\SumN \frac{\xi_t^2}{\phi_{0:t-1}} \notag \\ 
		&\leq 2\left(\sigma D^2 \!+\! \frac{L_f}{\sigma}\right) \sqrt{h_{1:T} +\mu_{T+1}}+\SumN  a_{t-1}\xi_t^2 \triangleq \widehat B_T \notag
	\end{align}
	Therefore similarly to \eqref{lambda-rel}, we can write:
	\begin{align*}
		&\|\ll_{T+1}\!\|\leq\! a_T\sqrt{	2 \big(\widehat B_T-\c R_T  \big)/a_{T} }+a_1G. 
	\end{align*}
	And finally, observe that since the dual regularizer $q_t(\ll)$ and learning rate $a_t$ are given by \eqref{dual-reg}, \eqref{dual-rate}, the bounds  \eqref{bound-eq3}, \eqref{bound-eq2} hold for LLP2 as well. Putting these together, we arrive at:
	\begin{align*}
		&\|\ll_T\|\!=\!\O \left( \max\left\{ T^{ \frac{k+1+2\theta}{4}}, T^{\frac{1+3\theta}{4} }, T^{ \frac{n+\theta}{2}}, T^{\frac{1+\theta}{2}} \right\} \right).
	\end{align*}
	That is, the growth rate of $\|\ll_T\|$ remains $\O(T^{(1+\theta)/2})$ and is not affected by the non-proximal primal regularizers. Similarly, the growth rate of $\widehat B_T$ is the same as that of $B_T$, see \eqref{eq4}, since $h_{1:T}\!=\O(T^{(3+\theta)/2})$ dominates $\mu_{T+1}\!=\!\O(T^{1+\theta})$. Therefore, the growth rate of $\c R_T$ and of $\c V_T$ are exactly as those of LLP. 
	
On the other hand, the $\c R_T$ bound of LLP2 is not zeroed for perfect predictions. Indeed, when $\epsilon_t=\delta_t=\xi_t=\!0, \forall t$, and $h_{1:t}=0$, therefore:
\begin{align}
	&\c R_T\leq \sqrt{\mu_{T+1}}\,\,\,\text{and}\,\,\,\,\c V_T\leq \sqrt{-2\c R_T/a_{T-1}} + \sqrt{\mu_{T+1}} \notag
\end{align}
where $\mu_{T+1}\!=\!E_m\!+\!a_{T}(T\!+1)G\Delta_m>0$ and $a_T=aT^{-\beta}$.  Hence, $\c R_T=\O(T^{(1-\beta)/2})$ and $\c V_T=\O\big(T^{(1+\beta)/2}\big)$ which shows that we cannot achieve zero regret, and even more so, that as we reduce the bound of $\c R_T$ by increasing $\beta$, we deteriorate in a commensurate amount the bound of $\c V_T$.

\subsection{Proof of Lemma \ref{lem:linearized} }

If we define $\L_t(x,\ll)$ as in \eqref{eq:lagrangian} but replace $g_t(x)$ with $g_t^{\ell}(x)$ and also use the linearized prediction $\p g_t^{\ell}(x)$, with:
\begin{align}
	&g_t^{\ell}(x)=g_t(x_t)+ \grad g_t(x_t)^\top (x-x_t), \notag \\
	 &\p g_t^{\ell}(x)=\p g_t(\p x_t)+ \grad \p g_t(\p x_t)^\top (x-\p x_t),  
\end{align}
then, the updates that we use in LLP2 are: 
\begin{align}
x_t&\!=\!\arg\min_{x\in \mathbb{R}^N} \Bigg\{ \sum_{i=0}^{t-1} \L_i(x, \ll_i) \!+ \p c_t^\top x \!+ \ll_t^\top \p g_t^{\ell}(x) \Bigg\} \notag \Rightarrow \\
x_t&\!=\! \arg\min_{x\in \mathbb{R}^N} \Bigg\{ r_{0:t-1}(x)+ c_{0:t-1}^\top x+ \sum_{i=0}^{t-1} 
\lambda_i^\top g_i^\ell(x) \notag \\
&+\p c_t^\top x+ \ll_t^\top \big(  \grad\p g_t(\p x_t)^\top x \big) \Bigg\}\notag \\
z_t&=\arg\min_{x\in \mathbb{R}^N} \left\{ \sum_{i=0}^{t} \L_i(x, \ll_i) \right\}\notag \\
&=\arg\min_{x\in \mathbb{R}^N} \left\{ r_{0:t}(x)  + c_{1:t}^\top x + \sum_{i=0}^t\ll_i^\top g_i^\ell(x)  \right\} \notag \\
&\!\!\!\lambda_{t+1}\!=\!\arg\max_{\ll\in\mathbb{R}^d} \left\{ \sum_{i=0}^t\L_i(z_i,\ll_i)+\lambda^\top \p g_{t+1}^{\ell}(\p x_{t+1})\right\} \label{dual-update2} \\
&\!=\!\arg\max_{\ll\in\mathbb{R}^d} \left\{ \lambda^\top\Big(\sum_{i=1}^t g_i^\ell(z_i)\!+ \p g_{t+1}(\p x_{t+1}) \Big)\! -\! \phi_{0:t}\|\ll\|^2/2  \right\} \notag
\end{align}
where the primal and dual regularizers are given again by \eqref{primal-reg}, \eqref{dual-reg} using the modified error parameters, $\forall t$:
\begin{align*}
&\delta_t=\grad g_t(x_t) - \grad \p g_t(\p x_t),\quad \xi_t=\| g_t^\ell(z_t)- \p g_t^\ell(\p x_t) \|,\quad \text{and} \\
&\epsilon_t=c_t-\p c_t \,\,\,\,\,\text{as before}.
\end{align*}
Note that in case of perfect predictions, i.e., when:
\[
\grad \p g_t(\p x_t)=\grad g_t(x_t), \quad c_t=\p c_t, \quad \p g_t(\p x_t)= g_t(x_t),\,\,\,\forall t,
\]
then $\epsilon_t=0$, $\delta_t=0$, and 
\begin{align*}
\xi_t&=\Big\| g_t(x_t)+\grad g_t(x_t)^\top (z_t-x_t) - \p g_t(\p x_t) \notag \\
&- \grad \p g_t(\p x_t)^\top (\p x_t-\p x_t)\Big\|=\left\| \grad g_t(x_t)^\top (z_t-x_t)\right \|=0
\end{align*}
where the last step follows as for perfect predictions, clearly, it holds $z_t=x_t$.

Next, it is easy to see that Lemma \ref{2ssclaim2b} holds and yields the same bound $\|x_t-z_t\|\leq \|\epsilon_t+\lambda_t^\top \dd_t\|/\sigma_{1:t}$ with the redefined $\{\delta_t\}$. Applying \cite[Theorem 2]{mohri-aistats2016} to \eqref{dual-update2}, we get:
\begin{align}
	&-\sum_{t=1}^T \ll_t^\top g_t^\ell(z_t)+ \sum_{t=1}^T \ll^\top g_t^\ell(z_t)\notag \\ &\leq q_{0:T-1}(\ll)+ \sum_{t=1}^T \|g_t^\ell(z_t)-\p g_t(\p x_t)\|_{(t-1),\star}^2= \notag \\
	&=q_{0:T-1}(\ll)+ \sum_{t=1}^T \frac{\|g_t^\ell(z_t)-\p g_t(\p x_t)\|^2}{\phi_{0:t-1}},\,\,\,\forall \ll\in\mathbb R_+^d. \label{eq:dual1b}
\end{align}
Then, we can repeat the analysis in Sec. \ref{sec:regret}, noting:
\[
g_t^\ell(x^\star)=g_t(x_t)+\grad g_t(x_t)^\top(x^\star - x_t)\preceq g_t(x^\star) \preceq 0, \,\forall t
\]
due to convexity of $g_t(\cdot)$ and the property of $x^\star$, to arrive at the same bound $B_T$ for the regret $\c R_T$ -- sans the redefined $\{\delta_t\}$ and $\{\xi_t\}$ parameters.

Similarly, repeating the analysis of Sec. \ref{sec:constraint} we get:
\begin{align}
\!\!\lambda^\top \left( \SumN g_t^\ell(z_t)\right)\!-\!\frac{\phi_{0:T-1}}{2}\|\ll\|^2 \leq B_T\!-\!\c R_T,\forall \ll\in\mathbb{R}_+^d. \label{eq3b}
\end{align}
Minimizing the LHS, similarly to Sec. \ref{sec:constraint} , we obtain:
\[
V_T^{z,\ell}\!\leq\! \sqrt{ 2(B_T-\c R_T)/a_{T-1}} \,\,\, \text{where} \,\,\, \c V_T^{z,\ell}\!\triangleq\!\left \| \left[\SumN \!g_t^\ell(z_t)	\right]_+ \right\|.
\]
Finally, note that using the definition of $g_t^\ell(x)$, we can write:
\begin{align*}
&g_t(x_t)=g_t^\ell(z_t)-\grad g_t(x_t)^\top (z_t-x_t ) \Rightarrow \notag \\
&\SumN g_t(x_t)=\SumN g_t^\ell(z_t)+\SumN\grad g_t(x_t)^\top (x_t-z_t ).
\end{align*}
Hence, we obtain the bound:
\begin{align*}
&\left\| \left[\SumN g_t(x_t)	\right]_+ \right\| =  \left\| \left[\SumN g_t^\ell(z_t)+\grad g_t(x_t)^\top (x_t-z_t )	\right]_+ \right\| \notag \\
&\leq \left\| \left[\SumN g_t^\ell(z_t)\right]_+\right\|+\left\|\left[\SumN \grad g_t(x_t)^\top (x_t-z_t )	\right]_+ \right\|,
\end{align*}
where we used the identity $\| [\chi +\upsilon]_+ \|\leq \|[\chi]_+\|+ \| [\upsilon]_+\|$. Therefore we arrived at the same result as with the non-linearized constraints:
\begin{align*}
\c V_T\leq V_T^{z, \ell}+\frac{2L_g}{\sigma}\sqrt{h_{1:T}} = \sqrt{\frac{2(B_T-\c R_T)}{a_{T-1}}} +\frac{2L_g}{\sigma}\sqrt{h_{1:T}},
\end{align*}
by using a slightly different proof.

Finally, it follows directly from the proof of Theorem \ref{Theorem2} that the convergence rates are not affected by this linearization of the constraints. We conclude by stressing that in this version of LLP we only required predictions $\p c_t$, $\grad \p g_t(\p x_t)$ and $\p g_{t+1}(\p x_{t+1})$; while with non-linearized constraints LLP was using predictions $\p c_t$, $\p g_t(\cdot)$ and $\p g_{t+1}(\p x_{t+1})$.

\subsection{Proof of Lemma \ref{lem:linear-const}} \label{appendix:linear-const}
For this specific type of constraints, the primal and prescient updates are as follows:
\begin{align*}
	&x_t=\arg\min_{x\in \c X} \Big\{r_{0:t-1}(x)+c_{1:t-1}^\top x + \p c_t^\top x + \lambda_{1:t}g(x) \Big\} \\
	&z_t=\arg\min_{x\in \c X} \Big\{r_{0:t}(x)+c_{1:t}^\top x + \lambda_{1:t}g(x) \Big\},
\end{align*}
while the dual update remains the same:
\begin{align*}
	\lambda_{t+1}\!=\arg\max_{ \lambda \in \mathbb R_+^d} &\Big\{\!-q_{0:t}(\lambda)\!+\!\lambda^\top \Big( \p g_{t+1}(\p x_{t+1})\!+\sum_{i=0}^t g_t(z_i)\Big)\Big\}.
\end{align*}

With this modification, Lemma \ref{2ssclaim2b} yields the bound:
\begin{align*}
	\|x_t-z_t\| \leq \frac{\|\epsilon_t\|}{\sigma_{1:t}},\,\,\,\,\text{with}\,\,\,\epsilon_t=c_t-\p c_t,\,\,\sigma_t=\sigma\sqrt{ \sum_{i=1}^t \|\epsilon_i\|}
\end{align*}
and therefore:
\begin{align*}
	&\sum_{t=1}^T c_t^\top (x_t-z_t)\leq \sum_{t=1}^T \frac{ \|c_t\|\|\epsilon_t\|}{\sigma_{1:t}} \leq 2\frac{L_f}{\sigma}\sqrt{ \sum_{t=1}^T \|\epsilon_t\|}.
\end{align*}
Similarly, we can redefine $B_T$ and re-derive \eqref{eq3} as follows:
\begin{align*}
	&\c R_T + \lambda^\top \left( \sum_{t=1}^T g_t(z_t)-\frac{\|\lambda\|^2}{2a_{T-1}}\right)\leq B_T\\
	&\text{where } B_T\triangleq 2\left(\sigma D^2 + \frac{L_f}{\sigma}	\right) \sqrt{ \sum_{t=1}^T \|\epsilon_t\| } + \sum_{t=1}^T a_{t-1}\xi_t^2
\end{align*}
and $\xi_t=\|g_t(z_t)-\p g_t(\p x_t)\|$. Selecting the $\lambda$ that maximizes the second term in the LHS as before, we arrive at:
\begin{align}
\!\!\!\c R_T \!+\! \frac{a_{T-1}}{2}\left( V_T^z	\right)^2\!  \leq \! B_T \,\,\text{and } \c V_T\leq V_T^z \!+\! \frac{2L_g}{\sigma}\sqrt{\sum_{t=1}^T \|\epsilon_t\| } \!\!\label{temp1}
\end{align}
From this result, we can see directly that when $\epsilon_t=\xi_t=0,\forall t$, we get $\c R_T\leq 0$ since $B_T=0$ and $a_{T-1}$ is positive. And for the constraints, it holds:
\begin{align*}
	V_T\leq \sqrt{ \frac{-2\c R_T}{a_{T-1}} }=\c O\left(T^{\frac{1+\beta}{2}}\right).
\end{align*}
For worst-case predictions, we can drop the positive term $a_{T-1}(V_T^z)^2$ and write:
\[
\c R_T\leq B_T = \c O\Big(\max\Big\{\sqrt T, T^{1-\beta} \Big\} \Big),
\] 
and replace $-\c R_T\leq FT=\c O(T)$ in \eqref{temp1} and rearrange to obtain the bound for $V_T^z$ and then using the relation of $\c V_T$ to $V_T$ (see \eqref{temp1}) to bound the former. Finally, it is interesting to observe that the derivation of the bounds did not require explicitly bounding the dual variables, and this stems from the fact that $x_t$ is independent of the constraint perturbations.

\begin{figure*}[h!]
	\begin{subfigure}{.195\textwidth}
		\centering
		\includegraphics[width=1.\linewidth]{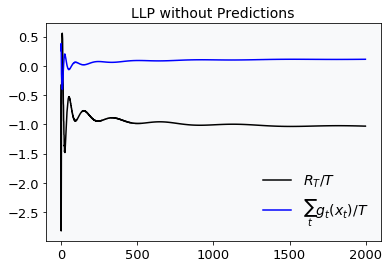}
		\label{fig:verification1b}
	\end{subfigure}
	\begin{subfigure}{.195\textwidth}
		\centering
		\includegraphics[width=1.\linewidth]{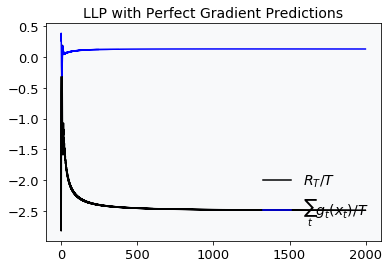}
		\label{fig:verification2b}
	\end{subfigure}
	\begin{subfigure}{.195\textwidth}
		\centering
		\includegraphics[width=1.\linewidth]{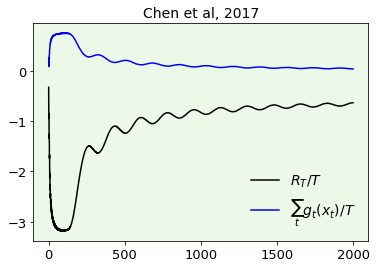}
		\label{fig:verification3b}
	\end{subfigure}
	\begin{subfigure}{.195\textwidth}
		\centering
		\includegraphics[width=1.\linewidth]{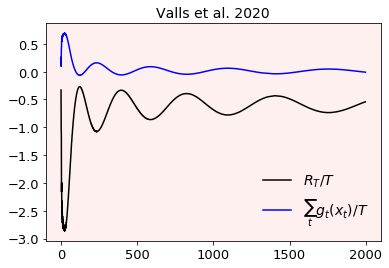}
		\label{fig:verification4b}
	\end{subfigure}
	\begin{subfigure}{.195\textwidth}
		\centering
		\includegraphics[width=1.\linewidth]{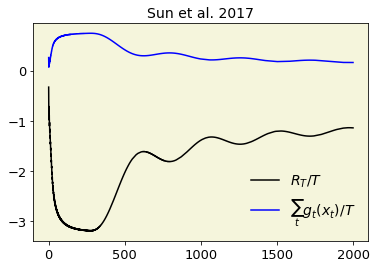}
		\label{fig:verification5b}
	\end{subfigure}
	\caption{{\small{Benchmark comparisons for: $f_t(x)\!=\!-4x$ if $t\% 2\!=\!0$, $f_t(x)\!=\!-x$ otherwise; $g_t(x)\!=\!0.79x\!+0.26$ if $t\%2\!=\!0$, $g_t(x)\!=\!0.64x\!-0.135$ otherwise; $\mathcal X\!=\![-1,1]$. The 1st LLP plot does not use any predictions, while the 2nd LLP plot uses perfect predictions for $\nabla f_{t}(x_{t})$ and $\nabla g_{t}(x_{t})$ {but no predictions} for $g_{t+1}(x_{t+1})$.} }}
	\label{fig:1}	
\end{figure*}


\begin{figure*}[t]
	\begin{subfigure}{.195\textwidth}
		\centering
		\includegraphics[width=1.\linewidth]{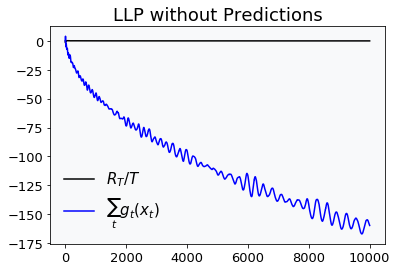}
	\end{subfigure}
	\begin{subfigure}{.195\textwidth}
		\centering
		\includegraphics[width=1.\linewidth]{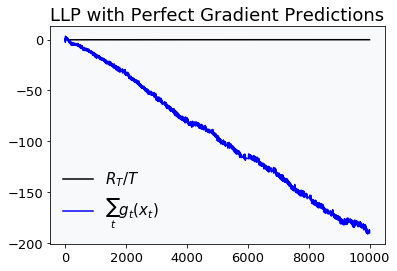}
	\end{subfigure}
	\begin{subfigure}{.195\textwidth}
		\centering
		\includegraphics[width=1.\linewidth]{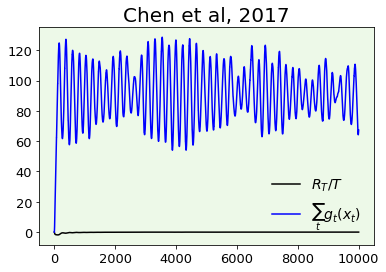}
	\end{subfigure}
	\begin{subfigure}{.195\textwidth}
		\centering
		\includegraphics[width=1.\linewidth]{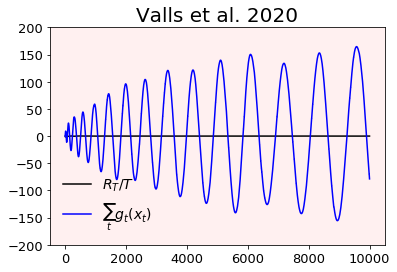}
	\end{subfigure}
	\begin{subfigure}{.195\textwidth}
		\centering
		\includegraphics[width=1.\linewidth]{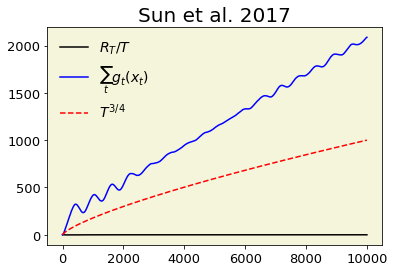}
	\end{subfigure}
	\caption{{\small{Benchmark comparisons for: $f_t(x)=-2x$; $g_t(x)=x$ with probability $0.1/(t+1)^{0.05}$ and $g_t(x)=-0.01$ otherwise; $\mathcal X\!=\![-1,1]$.} } } 
\end{figure*}

\subsection{Numerical Tests}

Finally, we conclude by providing some simple, yet illuminating, numerical results comparing LLP with three competitor algorithms: the MOSP algorithm by Chen et al. \cite{giannakis-TSP17}; our previous work Valls et al. \cite{victor}; and Sun et al. \cite{kapoor-icml17}.

Figure \ref{fig:1} presents the first set of results. The algorithms run on the following cost and constraint functions:
\[
f_t(x)=\begin{cases} -4x &\text{ if } t \text{ mod } 2=0 \\
	-x & \text{ otherwise }  \end{cases}.
\]
and
\[
g_t(x)=\begin{cases} 0.79x+0.26 &\text{ if } t \text{ mod } 2=0 \\
	0.64x-0.135 & \text{ otherwise }  \end{cases}.
\]
where $x\in \mathcal X=[-1,1]$. For the first LLP run we do not use any predictions, so as to demonstrate the efficacy of the algorithm even when no predictions are available. The second LLP plot runs the linearized version of the algorithm and uses perfect gradient predictions for the cost and constraint functions, but no predictions for the next constraint value. The three competitors have been optimized for $\c R_T$, by using the steps and tuning parameters that are suggested in their respective references. We observe that LLP achieves lower regret from all competitors, and it reaches that point faster.

In the second experiment, we run the algorithms on the time-invariant cost function $f_t(x)=-2x$, $\forall t$, and constraint function:
\[
g_t(x)=\begin{cases} x &\text{ with probability } \frac{0.1}{(t+1)^{0.05}} \\
-0.01 & \text{ otherwise }  \end{cases}.
\]
where, again, $x\in \c X=[-1,1]$. Note that in this example we plot the total, not the average, constraint violation so as to shed light on the actual operation of each algorithm.We observe that LLP satisfies continuously the constraints in each $t$, while the competitors oscillate or fail to converge, despite that the cost function is constant.

The above results demonstrate that LLP performs quite well in practice, where even in these simple examples (one dimension space, time-invariant cost functions, etc.) it has clear advantages over the state-of-art competitors. For example, we see that it achieves fast lower regret points (first experiment); and is able to handle the probabilistic constraints in the second example -- which is not surprising given that it uses a lazy dual update scheme which turns to be robust in such variations.

\end{document}